\documentclass{article}

\usepackage[preprint,nonatbib]{neurips_2026}

\usepackage[utf8]{inputenc}
\usepackage[T1]{fontenc}
\usepackage{hyperref}
\usepackage{url}
\usepackage{booktabs}
\usepackage{amsfonts}
\usepackage{nicefrac}
\usepackage{microtype}
\usepackage{xcolor}

\usepackage{graphicx}
\usepackage{amsmath}
\usepackage{amssymb}
\usepackage{amsthm}
\usepackage{mathtools}
\usepackage{multirow}
\usepackage{makecell}
\usepackage{array}
\usepackage{longtable}
\usepackage{enumitem}
\usepackage{caption}
\usepackage{subcaption}
\usepackage[many]{tcolorbox}
\usepackage{cleveref}
\usepackage[numbers,sort,compress,sectionbib]{natbib}

\usepackage{times}
\usepackage{epsfig}
\usepackage{bbding}
\usepackage{wrapfig}
\usepackage{algorithm, algpseudocode}

\theoremstyle{plain}

\theoremstyle{definition}

\theoremstyle{remark}

\title{HiMA-Ecom: A Long-Horizon Memory Benchmark for Hierarchical Multi-Agent E-commerce Assistants}

\newcommand{\blfootnote}[1]{%
  \begingroup
  \renewcommand{\thefootnote}{}%
  \footnotetext{#1}%
  \endgroup
}

\author{%
  \bfseries Junxing Hu\textsuperscript{1,*} \hspace{0.7em}
  Ai Han\textsuperscript{1,*,\dag} \hspace{0.7em}
  Haolan Zhan\textsuperscript{2} \hspace{0.7em}
  Pu Wei\textsuperscript{1} \hspace{0.7em}
  Zhiqian Zhang\textsuperscript{1} \hspace{0.7em}
  Yuhang Guo\textsuperscript{1} \\
  \bfseries Jiawei Lu\textsuperscript{1} \hspace{0.7em}
  Zhen Chen\textsuperscript{1} \hspace{0.7em}
  Haoran Li\textsuperscript{3} \hspace{0.7em}
  Zicheng Zhang\textsuperscript{1} \hspace{0.7em}
  Ling Li\textsuperscript{4} \\[0.6em]
  \normalfont \textsuperscript{1}JD.com \qquad \textsuperscript{2}Monash University \\
  \normalfont \textsuperscript{3}University of the Chinese Academy of Sciences \qquad \textsuperscript{4}Microsoft
}

\begin{document}

\maketitle

\blfootnote{\textsuperscript{*}Equal contribution. \textsuperscript{\dag}Corresponding author.}

\begin{abstract}
Long-horizon agentic systems become long-context systems by accumulation rather than through a single long document, since every step reassembles a prompt from role instructions, tool and sub-agent descriptions, recalled memories, dialogue history, and previous tool outputs, separately for every agent.
This paper studies how such accumulated context can be trained, evaluated, and managed in hierarchical multi-agent assistants.
We present HiMA-Ecom, a hierarchical multi-agent benchmark for e-commerce assistants with 22.8K instances, of which 17.7K carry memory entries in context, pairing agent-specific supervised fine-tuning data with system-level supervision for end-to-end multi-agent reinforcement learning.
Building upon it, HiMA-R1 keeps the context budget of a trajectory under control through Variance-Reduction Group Relative Policy Optimization (VR-GRPO), which samples along an initial trajectory so that sampling cost grows additively with trajectory length and updates the agents with the largest reward variance, together with an efficiency reward on trajectory length and a memory evolution mechanism that repurposes GRPO rewards as cost-free signals for what stays in context.
Agents built upon 3B and 7B open-source models match DeepSeek-R1 and surpass DeepSeek-V3 by 6.0 points on average, while reward-driven memory reduces the update steps needed for convergence by roughly 55\%.
\end{abstract}

\section{Introduction}

The rapid advancement of LLMs~\cite{achiam2023gpt, AnthropicClaude3, grattafiori2024llama, bai2023qwen, yang2024qwen2} has empowered agents to reason, plan, and call tools over long horizons~\cite{guo2024large, gao2025survey}.
Compared to single agents with specialized functionalities~\cite{li2023api, ruan2023tptu, qin2024toolllm, dong2024survey}, multi-agent systems~\cite{wan2025rema, liao2025marft, dang2025multi} are more flexible on complicated tasks such as AI assistants~\cite{fu2024camphor} and emergency response~\cite{sivagnanam2024multi}, where Multi-Agent Reinforcement Learning (MARL) is a key technique~\cite{ning2024survey}.

What makes such assistants long-context is accumulation rather than a single long input.
At every step, an agent is prompted with its role instructions, the descriptions of the tools and sub-agents it may invoke, the memories recalled for the current query, the dialogue history, and the outputs of previous tool calls.
This context is reassembled at every step and for every agent, so the prompt a system consumes grows with trajectory length and with the number of agents, several times beyond what a single agent needs for the same query.
The question is then not only how long a context window can be, but which parts of it are worth keeping and how a system learns to manage them.

\begin{figure}[t]
\centering
\includegraphics[width=1.0\textwidth]{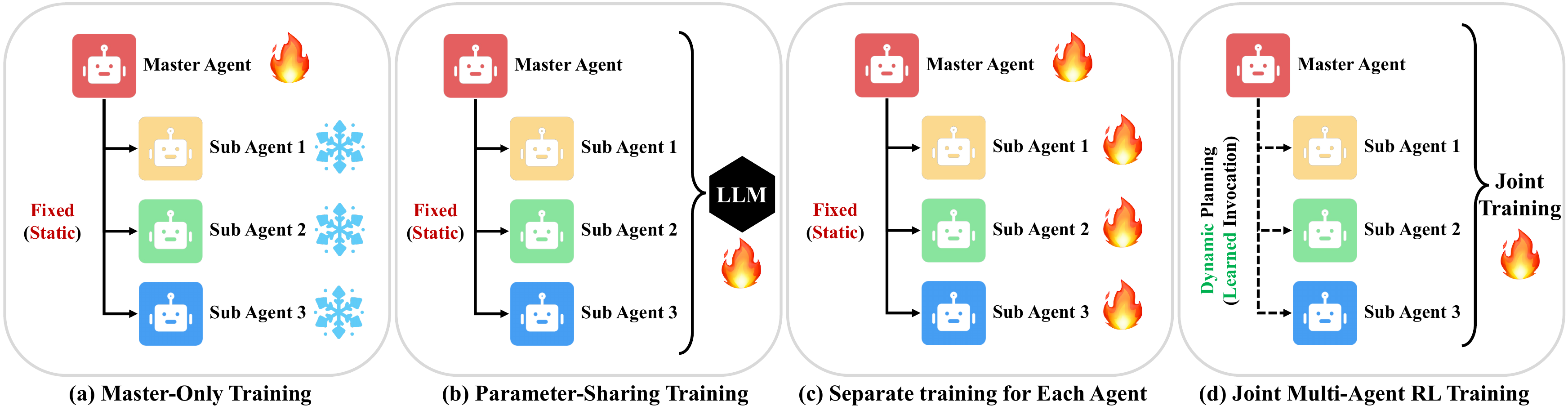}
\caption{
Comparison of training paradigms for hierarchical LLM-based multi-agent systems. (a) Master-Only Training, (b) Parameter-Sharing Training, (c) Separate Training with fixed planning, and (d) Joint Multi-Agent RL Training with dynamic planning. The first three adopt static invocation structures, while this work focuses on the joint training paradigm with learned invocation in (d),
}
\vspace{-3mm}
\label{training_comp_pic}
\end{figure}

As LLM-based agents are deployed in vertical domains such as e-commerce and customer service~\cite{fu2024camphor}, practical systems are often organized in a hierarchical master-sub-agent structure, where a master agent orchestrates and specialized sub-agents execute domain-specific subtasks.
As shown in Figure~\ref{training_comp_pic}, existing training paradigms rely on static invocation structures such as master-only optimization, parameter sharing, and isolated agent training with fixed planning.
They are limited when agent roles, interaction patterns, and reasoning strategies must co-evolve to handle long-horizon requests, which motivates the joint paradigm in Figure~\ref{training_comp_pic}(d) with learned invocation.

Since RL is effective for aligning models with human preferences~\cite{ouyang2022training}, LLM-based MARL methods have flourished in task decomposition~\cite{iqbal2022alma, tian2023decompose} and adaptive coordination~\cite{fu2022revisiting, li2024language}, although the behavior of one agent affects the rewards of others~\cite{hernandez2017survey}.
Multi-Agent Proximal Policy Optimization (MAPPO)~\cite{yu2022surprising} adapts PPO~\cite{schulman2017proximal} to multi-agent settings but relies on value functions, whose decoupled updates often destabilize training across heterogeneous agents with misaligned rewards~\cite{zhong2024heterogeneous}.
Group Relative Policy Optimization (GRPO)~\cite{shao2024deepseekmath}, popularized by DeepSeek-R1~\cite{guo2025deepseek}, replaces critics with population-based comparisons and performs well for single or homogeneous agents~\cite{deng2024smac, xia2025gui, lu2025ui} and vision-language models~\cite{huang2025vision}.
Applying it to functionally distinct agents is harder, since sampling paths grow multiplicatively with trajectory length, each agent carries its own accumulated context, and one end-to-end reward is shared across different roles.
Benchmarks reflecting these structural, memory, and coordination demands remain scarce.

To bridge this gap, we introduce HiMA-Ecom, a hierarchical multi-agent benchmark for e-commerce assistants that supports agent-level Supervised Fine-Tuning (SFT) with memory as well as system-level multi-agent RL, together with HiMA-R1, a joint training framework in which sampling, rewards, and memory all act on the context that agents accumulate along a trajectory.
The main contributions of this work are listed as follows:

\begin{itemize}[leftmargin=*]

\item We introduce HiMA-Ecom, to our knowledge the first hierarchical multi-agent benchmark that pairs agent-specific memory-augmented SFT data with system-level RL supervision, where 17.7K of its 22.8K instances carry memory entries in context and no intermediate agent-level supervision is observable.

\item We describe HiMA-R1, a joint training framework that keeps the context budget of a trajectory under control through additive Monte Carlo sampling, variance-based selective updates, an efficiency reward, and reward-driven memory evolution.

\item Agents built upon 3B and 7B open-source models become comparable to substantially larger LLMs on the benchmark, memory reduces the update steps to convergence by roughly 55\%, and we account for the token cost of the accumulated context.

\end{itemize}

\section{Related Work}

LLM-based agents plan through structured prompts and self-reflection~\cite{yao2023react, shinn2023reflexion, qiao2024autoact}, and multi-agent systems extend them to settings that need distinct capabilities~\cite{kannan2024smart, shen2024small, fu2024camphor}, including orchestration over agent graphs~\cite{zhuge2024gptswarm, zhang2024aflow} and debate among peers~\cite{du2024improving}.
Cooperative MARL is usually built on actor-critic frameworks such as MAPPO~\cite{yu2022surprising}, whereas GRPO~\cite{shao2024deepseekmath} removes the value function and therefore suits heterogeneous agents~\cite{ke2025survey}, with MLPO~\cite{estornell2025train} training the leader of a system and ReMA~\cite{wan2025rema} and MARFT~\cite{liao2025marft} training agents on reasoning-style tasks.
For memory, RAG-based mechanisms place recalled entries directly into the prompt, where selection~\cite{li2023mot, liu2023think, kagaya2024rap}, forgetting~\cite{zhong2024memorybank}, and aggregation~\cite{sarch2023open, zhao2024expel, xu2025mem} decide what an agent reads.
ReMA pairs a meta-thinking agent with a reasoning agent on tasks without tools or memory, while MemoryBank and ExpeL decide what to keep through a forgetting curve or LLM reflection. The setting studied here differs in that agents are functionally distinct, each carries its own memory-augmented context, and only the final system response is supervised, so memory competes for the context budget of every step and is optimized by the same reward that trains the policies.
A fuller discussion is provided in Appendix~\ref{Related}.

\section{The HiMA-Ecom Benchmark}\label{benchmark}

HiMA-Ecom is a hierarchical multi-agent benchmark for training and evaluating LLM-based assistants in vertical domains, with a focus on e-commerce.
It contains 22.8K instances, 17.7K of which carry memory entries inside the agent context, and supports both agent-specific supervised fine-tuning (SFT) and system-level reinforcement learning (RL) over a master-sub-agent architecture.

\begin{figure}[t]
\centering
\includegraphics[width=0.78\textwidth]{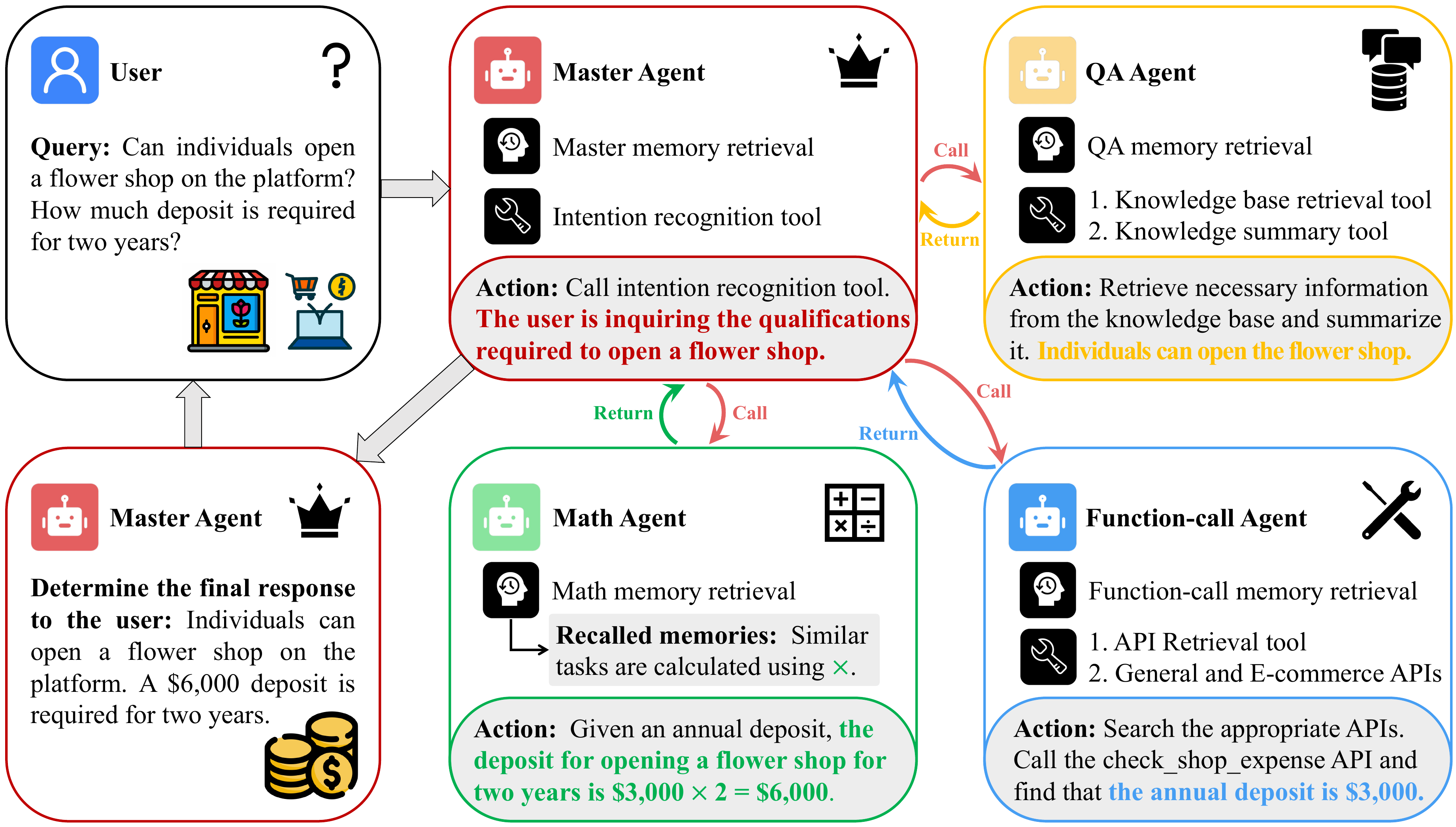}
\caption{
An illustrative example of the hierarchical multi-agent system under e-commerce scenarios, where each agent re-assembles its own context from role instructions, tool and sub-agent descriptions, recalled memories, dialogue history, and tool responses. Agent-level reasoning is visualized for clarity, while the dataset provides agent-specific SFT entries and joint input and output pairs for RL.
}
\vspace{-3mm}
\label{hima_pic}
\end{figure}

\vspace{-2mm}
\subsection{Hierarchical multi-agent architecture}
\vspace{-2mm}

As illustrated in Figure~\ref{hima_pic}, the system consists of one master agent and four functionally distinct sub-agents.
The \textbf{master agent} analyzes the query, orchestrates sub-agents or tools such as intention recognition at each step, and determines the final response.
The \textbf{question-answering agent} retrieves and summarizes information from external databases, the two \textbf{function-call agents} cover general and e-commerce APIs and either call an API directly from memory-driven prompts or query a tool retriever first, and the \textbf{math agent} solves mathematical problems.
Each agent is independently parameterized without parameter sharing and keeps its own memory pool.

\vspace{-2mm}
\subsection{What accumulates in the context of an agent}\label{context_comp}
\vspace{-2mm}

\begin{figure}[t]
\centering
\includegraphics[width=1.0\textwidth]{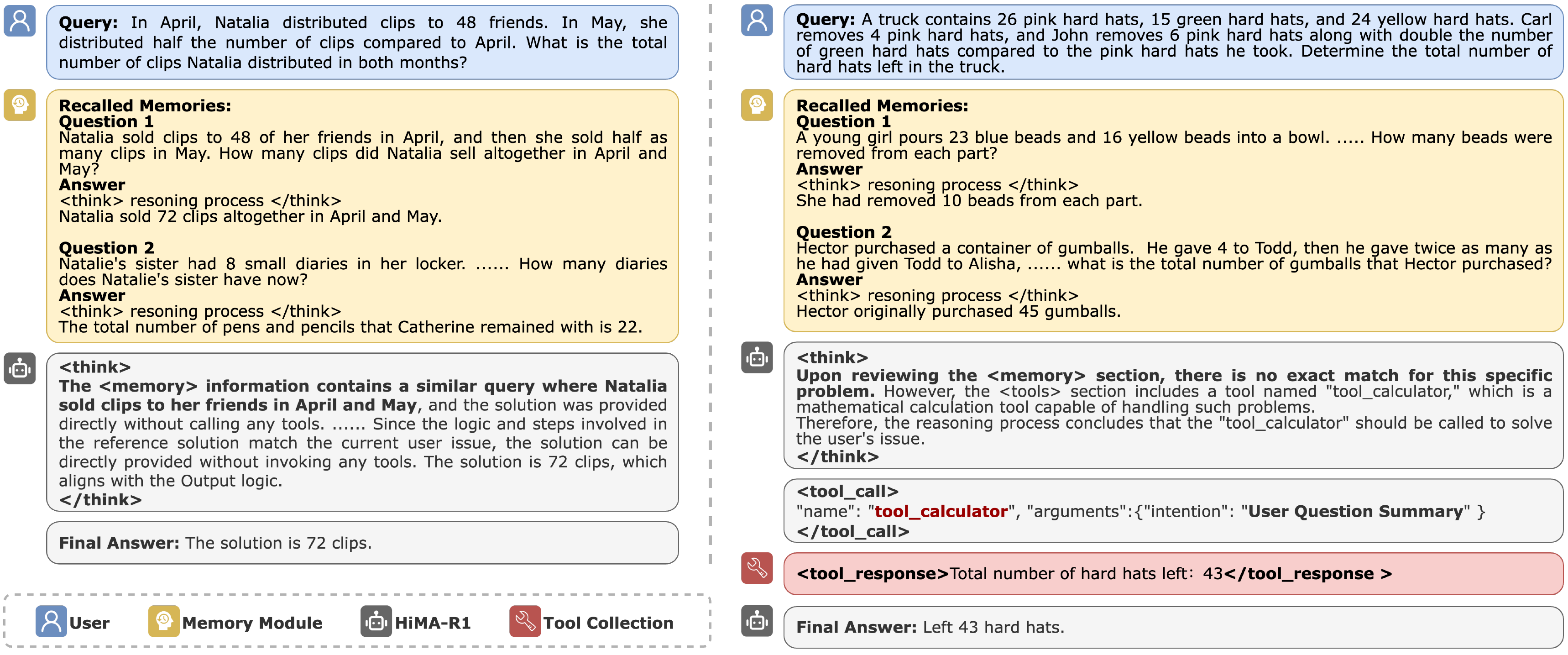}
\caption{
Examples of different agents making decisions based on dynamically recalled memories.
}
\vspace{-3mm}
\label{case_analysis}
\end{figure}

Every step is a fresh prompt assembled from five parts, which is what makes this setting long-context in practice.
The first is the role instructions, a per-agent template describing the role and the output protocol.
The second is the tool and sub-agent descriptions inside a \texttt{<tools>} block, listing four sub-agents plus retrieval and intention tools for the master, 12 e-commerce APIs for the e-commerce agent, and verbose ToolBench documentation~\cite{qin2024toolllm} for the general one.
The third is the recalled memories inside a \texttt{<memory>} block, where the top-5 entries retrieved for the query are placed, each storing a previous query, its plan, its answer, a reward-based score, and a timestamp.
The last two are the dialogue history and the tool responses appended verbatim, with prompts and examples in Appendix~\ref{Dataset} and Appendix~\ref{Prompt}.
Two properties follow. The prompt is rebuilt at every step and for every agent, so the token cost of one query grows with trajectory length and with the number of agents (Section~\ref{context_cost}).
The memory block is also the only part the system itself controls, since tools come from the environment and history from the trajectory, whereas what is recalled, kept, or discarded can be learned, which Section~\ref{grpomemory} optimizes.
Figure~\ref{case_analysis} shows both sides on training examples. A matching entry lets an agent reuse a stored answer, while an agent with no useful entry pays for the recalled context and still falls back to a tool call.

\vspace{-3mm}
\begin{table}[t]
\centering
\scriptsize
\setlength{\tabcolsep}{5pt}
\caption{Overview of the HiMA-Ecom benchmark.}
\begin{tabular}{lcccc}
\toprule
Agent & Domain & Training stage & Instances & Instances with memory \\
\midrule
Master & General + E-commerce & SFT & 13K & 9K \\
Question answering & General + E-commerce & SFT & 1K & 1K \\
E-commerce function call & E-commerce & SFT & 2K & 2K \\
General function call & General & SFT & 3.3K & 3.3K  \\
Mathematics & General & SFT & 2.4K & 2.4K \\
Collaboration & General + E-commerce & RL & 1.1K & - \\
\midrule
Total & General + E-commerce & SFT + RL & 22.8K & 17.7K \\
\bottomrule
\end{tabular}
\label{hima_data}
\vspace{-3mm}
\end{table}

\subsection{Construction and statistics}
\vspace{-2mm}

Table~\ref{hima_data} summarizes the composition of HiMA-Ecom.
\textbf{Supervised fine-tuning data} is agent-specific and teaches each agent both its domain skill and the use of memory.
The master data covers single sub-agent invocations and collaborative cases, the e-commerce function-call data covers 12 prevalent platform APIs, the general one is drawn from ToolBench~\cite{qin2024toolllm}, question answering combines real e-commerce queries with open-domain instances from COIG~\cite{zhang2023chinese}, and mathematics comes from GSM8K~\cite{cobbe2021training}.
Most instances include memory, drafted with GPT-4o~\cite{achiam2023gpt} and then manually modified, and supervision targets are reasoning processes, tool invocation decisions, and final outputs depending on the role. No personally identifiable or sensitive private information is included.

\textbf{Reinforcement learning data} contains only the initial user query and the final system response, without intermediate agent-level annotation.
The training set has 600 samples, 100 for each independent sub-agent task covering math, QA, general and e-commerce function calls, plus 200 collaborative instances, and the test set has 100 samples each for math, QA and e-commerce function calls, a larger ToolBench-derived split for general function calls, and 100 collaborative instances handcrafted to require at least two sub-agents, so credit assignment is system-level as in deployment.

\textbf{Difference from existing datasets.}
Long-context agent benchmarks usually stress a single long input or a long dialogue, whereas HiMA-Ecom stresses context that a hierarchical system assembles for itself in a vertical domain, since its memories, 12 platform APIs, and collaborative queries all come from e-commerce operations where repeated requests make memory reuse realistic (Appendix~\ref{Dataset}).

\section{The HiMA-R1 Training Framework}\label{method}
Based on HiMA-Ecom, HiMA-R1 is a joint training framework whose components each act on a different cost of a long trajectory.
Sampling along an initial trajectory and variance-based selective updates keep exploration and updating additive rather than multiplicative in trajectory length (Section~\ref{grposampling}), while an efficiency reward and reward-driven memory evolution bound how much context a query accumulates (Section~\ref{grpomemory}).

\begin{figure}[t]
\centering
\includegraphics[width=1.0\textwidth]{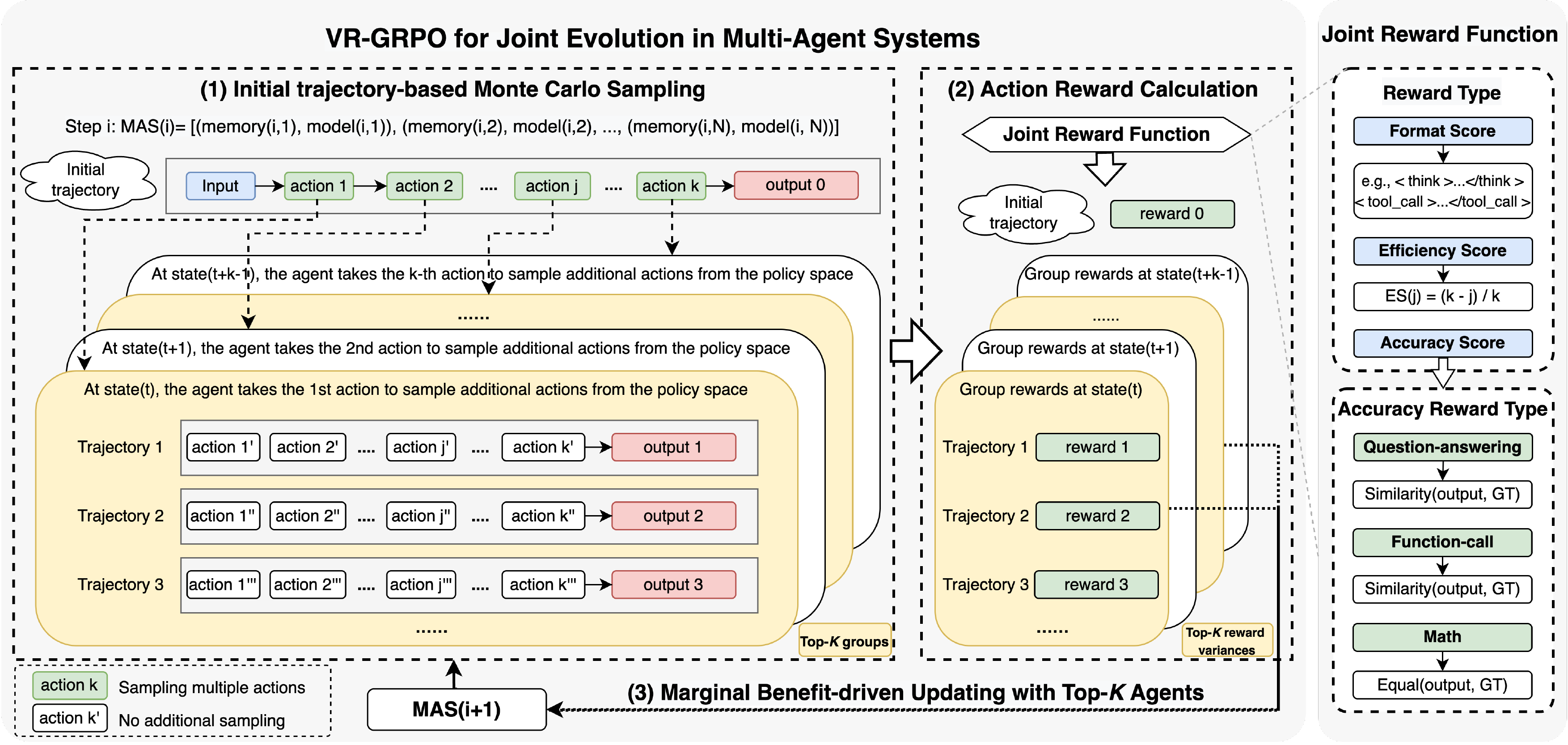}
\caption{
VR-GRPO for Multi-Agent Systems (MAS). (1) Initial trajectory-based Monte Carlo sampling sequentially samples agent actions along an initial trajectory, mitigating action space explosion. (2) Action reward calculation incorporates accuracy, format, and efficiency rewards based on the trajectory output. (3) Marginal benefit-driven updating prioritizes the top-$K$ agents with the highest reward variance to accelerate policy evolution.
}
\vspace{-3mm}
\label{dynamics}
\end{figure}

\subsection{Variance-Reduction GRPO}\label{grposampling}
Since actor-critic frameworks with critic models are inherently complex, HiMA-R1 employs GRPO for policy optimization, which foregoes the value function and computes advantages in a group-relative manner~\cite{shao2024deepseekmath}.
Tailored for multi-agent systems, Variance-Reduction Group Relative Policy Optimization (VR-GRPO) includes the following parts. Throughout, an \emph{agent} denotes a role together with its model, and a \emph{node} denotes one decision step of an agent along a trajectory, so the same agent may occupy several nodes.

\textbf{Initial trajectory-based Monte Carlo sampling.}
For a multi-agent system consisting of $N$ agents $\{ m_i\}_{i=1}^N$, each agent $m_i$ has an action space including $G_i$ possible actions.
When GRPO is applied naively, a reasoning trajectory of length $k$ generates $G_{Naive} = G_1 \times G_2 \cdots \times G_k$ sampling paths in an exponential explosion (Appendix~\ref{Paradigm}, Figure~\ref{sampling}(a)).
Instead, given a query $q$, VR-GRPO first produces an initial trajectory of length $k$, and then sequentially samples $(G_i-1)$ actions for each node $m_i$ in the trajectory, as shown in Figure~\ref{dynamics}(1).
The path from the query to a sampling node $m_j$ remains along the original trajectory, and after $m_j$ no more sampling is conducted while reasoning continues to the end, so a total of $G_{NMC} = G_1 + G_2 \cdots + G_k$ trajectories are generated.
This addition is much smaller than the multiplication, which matters most when trajectories are long and each node already carries a large context. Branching from one trajectory also gives every node a group of samples that share the same prefix, so their rewards are compared at that node, which independently sampled full trajectories do not provide.

\textbf{Marginal benefit-driven updating.}
For the $k$ groups of sampled trajectories, the parameters of the corresponding agents $\{m_i\}_{i=1}^k$ are $\Theta = \{\theta_i\}_{i=1}^k$, and each agent $m_i$ is optimized with the clipped group-relative objective $\mathcal{J}_{GRPO}(\theta_i)$ of Eq.~\ref{eq_GRPO_single} in Appendix~\ref{Objective}, where advantages are obtained by normalizing group-wise rewards.
The standard KL penalty is eliminated for greater policy divergence during coordination, without the overhead of maintaining reference models.
Updating every agent corresponds to $\mathcal{J}_{Multi\text{-}GRPO}(\Theta) = \{\mathcal{J}_{GRPO}(\theta_i) \mid i = 1, \cdots ,k \}$, which spends optimization on every node.
Instead, VR-GRPO restricts the update to the nodes whose rewards fluctuate most, following the marginal benefit principle~\cite{kauder2015history},
\begin{equation}
\begin{split}
\mathcal{J}_{VR-GRPO}(\Theta) =  \{\mathcal{J}_{GRPO}(\theta_i) \mid i \in \text{argtopK}\left( \text{Var}(R_i), K \right) \},
\end{split}
\label{eq_GRPO_V}
\end{equation}
where $R_i = \{r_j\}_{j=1}^{G_i}$ is the reward set obtained after sampling $G_i$ actions from $m_i$ to the end of trajectory, and $\text{argtopK}(\cdot)$ returns the index set of the $K$ nodes with the largest reward variance.
Reward variance is a cheap proxy for how much a node still has to gain, and it is available from the samples that have already been drawn, so the selection adds no extra cost to training. When the same agent is selected at several nodes, it is updated once per node, in descending order of reward variance, and the name of the method refers to this variance-based selection.

\textbf{Action rewards.}
In Figure~\ref{dynamics}(2), given the final answer $a$ corresponding to $q$ as the ground truth, the reward $\mathcal{R}$, denoted $r_i$ above, of each agent action consists of three terms,
\begin{equation}\label{eq_total_reward}
\mathcal{R}  = \mathcal{R_{A}} + \mathcal{R_{F}} - \mathcal{R_{E}}.
\end{equation}
The \textit{accuracy reward} ($\mathcal{R_{A}}$) is calculated end-to-end from the final answer, using semantic similarity for question answering and function calling and exact-match validation for mathematics.
The \textit{format reward} ($\mathcal{R_{F}}$) rewards the ReAct-style protocol, where thinking and tool calling are wrapped in $<\boldsymbol{think}>\cdots<\boldsymbol{/think}>$ and $<\boldsymbol{tool\_call}>\cdots<\boldsymbol{/tool\_call}>$ tags, which keeps outputs parsable when fed back into the context of another agent.
The \textit{efficiency reward} ($\mathcal{R_{E}}$) is computed for the $j$-th node in a trajectory of length $k$ as $\mathcal{R_{E}} = \frac{k-j}{k}$, imposing a penalty proportional to the distance of the node from the trajectory endpoint. Since every additional decision step appends another round of tool responses and history to the context of every downstream agent, this term prices the downstream context that an early plan commits the system to.

\subsection{Reward-driven memory evolution}\label{grpomemory}

Memory reduces redundant reasoning, but a pool that only grows becomes an expensive part of the prompt.
HiMA-R1 therefore lets memory evolve during training, using GRPO rewards as a cost-free supervisory signal instead of a dedicated model.

\textbf{Adaptive reward thresholding.}
The memory of each agent is updated with a unified reward that excludes the efficiency score, that is $\mathcal{R_{M}} = \mathcal{R_{A}} + \mathcal{R_{F}}$, so independent memory modules judge overall performance rather than their own position in the trajectory. For $G_{NMC}$ trajectories sampled from query $q$, the mean $\mu$ and standard deviation $\sigma$ of the corresponding rewards are computed, and the 2.5\% and 97.5\% percentiles of the approximate normal distribution are used as the lower bound $L = \mu - 1.96 \sigma$ and the upper bound $U = \mu + 1.96 \sigma$.

\textbf{Dynamic memory updating.}
The algorithm first decides whether to add a new memory based on the upper bound $U$, then updates the memories recalled by the query $q$ according to their similarity to the trajectory output and the trajectory reward, while all other memories decay with time. The full procedure is given in Appendix~\ref{Memalg}, Algorithm~\ref{alg_memory}.

\textbf{Memory overflow handling.}
To bound both storage and prompt length, a memory is deleted when its final reward $\mathcal{R_M}_{i}$ falls below a predefined threshold $D$, or when the pool exceeds its capacity and the reward rank of the memory is relatively low.

In this way, the content recalled into the context is curated by the same reward that trains the policies, which accelerates convergence and reduces repeated reasoning, as quantified in Section~\ref{EXPERIMENT}.

\section{Experiments}\label{EXPERIMENT}

\subsection{Implementation details}
We opt for the Qwen2.5 series~\cite{yang2024qwen2} as the agent backbone. Each agent is a separate model whose full parameter set is updated in both stages, without adapters or parameter sharing, so a query activates from 6B (master plus one sub-agent) to 15B (master plus four).
Base models are first fine-tuned with a learning rate of 5e-6 for 5 epochs, and the system is then trained with RL at 1e-6 for 5 epochs on 8 NVIDIA H200 GPUs, with the best results reported. More settings are provided in Appendix~\ref{Implementation}.
Accuracy uses judged semantic similarity to the reference answer for question answering and general function calling, exact numerical matching for mathematics, and the predicted API name for e-commerce function calling, while efficiency is the average number of reasoning rounds per query. The memory pool of each agent starts empty and is accumulated during RL training, and evaluation recalls from the final pool without updating it, so no test answer enters memory.

\begin{table}[!t]
\centering
\scriptsize
\setlength{\tabcolsep}{3pt}
\caption{
Comparisons on the HiMA-Ecom benchmark (\%). Under the open-source setting, bold is the optimal and `\_' is the suboptimal value. Closed-source methods ($*$) are only used for reference.
}
\begin{tabular}{lcccccc}
\toprule
Model & Math & QA & E-commerce FC & General FC & Collaboration & \textbf{Average} \\
\midrule
Claude3.5-sonnet $*$  & 98.0 & 38.0 & {1.0} & {62.5} & 2.0 & {40.3}\\
GPT‑4o mini $*$  & 47.0 & 31.0 & {0.0} & {65.5} & 0.0 & {28.7}\\
GPT‑4o $*$ & 85.0 & 35.0 & {56.0} & {83.0} & 6.0 & {53.0}\\
\midrule
DeepSeek-R1  & \textbf{98.0} & \underline{37.0} & 24.0 & 72.0 & \underline{7.0} & \textbf{47.6}\\
DeepSeek-V3  & \underline{96.0} & 32.0 & 10.0 & 68.0 & 2.0 & 41.6\\
Qwen2.5-32B  & 72.0 & \textbf{38.0} & 3.0 & 72.0 & 2.0 & 37.4\\
Qwen2.5-14B  & 80.0 & 29.0 & 0.0 & 42.0 & 1.0 & 30.4\\
Multiagent Debate (14B, 3 roles) & 95.0 & 33.0 & 0.0 & 40.6 & 1.0 & 33.9\\
Qwen2.5-3B (single agent, RL) & 56.0 & 9.0 & 0.0 & 58.5 & 1.0 & 24.9 \\
Qwen2.5-7B (single agent, RL) & 81.0 & 11.0 & 0.0 & 67.2 & 1.0 & 32.0 \\
Qwen2.5-14B (single agent, RL) & 82.0 & 27.0 & 2.0 & 63.6 & 1.0 & 35.1 \\
HiMA-SFT (7B master + 3B subs) & 65.0 & 13.0 & 42.0 & 73.0 & 3.0 & 39.2\\
\textbf{HiMA-R1} (3B master + 3B subs) & 68.0 & 22.0 & \textbf{48.0} & \textbf{76.0} & 6.0 & \underline{44.0} \\
\textbf{HiMA-R1} (7B master + 3B subs) & 75.0 & 35.0 & \underline{46.0} & \underline{73.1} & \textbf{9.0} & \textbf{47.6} \\
\bottomrule
\end{tabular}
\label{compare_result}
\vspace{-3mm}
\end{table}

\subsection{Comparisons on the HiMA-Ecom benchmark}

All baselines run under the same prompts as single agents with ReAct-based multi-turn interactions, Multiagent Debate~\cite{du2024improving} uses Qwen2.5-14B expanded into three roles, and Qwen2.5-3B/7B/14B are additionally trained with GRPO on the same RL data (Appendix~\ref{Prompt} and Appendix~\ref{comparative_results}). Table~\ref{compare_result} reports the results.
While the closed-source GPT-4o remains the strongest overall, HiMA-R1 with a 7B master and 3B sub-agents reaches an average accuracy of 47.6\%, matching DeepSeek-R1 and surpassing DeepSeek-V3 at 41.6\%, and both HiMA-R1 configurations dominate the specialized e-commerce function-call task with 48.0\% and 46.0\% against 24.0\% for DeepSeek-R1 and 10.0\% for DeepSeek-V3.
It also exceeds the entire Qwen2.5 series on average, including the 32B variant, and outperforms Multiagent Debate at 33.9\% built upon a larger backbone.
Under comparable activated parameters, joint multi-agent RL reaching 44.0\% with a 3B master and four 3B sub-agents is clearly better than a single Qwen2.5-14B trained with RL at 35.1\%, and joint training lifts the math and QA scores of the SFT variant from 65.0\% and 13.0\% to 75.0\% and 35.0\%.
Collaboration remains the bottleneck for every system, with the best configuration at 9.0\%, far from saturated.

\textbf{Transfer to ToolBench.}
The same system is evaluated on ToolBench~\cite{qin2024toolllm}, whose test set is split by whether the tools were seen during training.
HiMA-R1 with 3B agents reaches 51.2\% in-domain and 57.5\% out-of-domain on average, above DeepSeek-R1 at 47.1\% and 51.8\% and DeepSeek-V3 at 43.1\% and 54.2\% with far fewer parameters, and it surpasses GPT-4o on the argument-level Hard\_F1 and F1 metrics in both settings. The full table, metrics, and baselines are in Appendix~\ref{comparative_results}.

\begin{table}[t]
\centering
\scriptsize
\begin{minipage}[t]{0.58\textwidth}
\centering
\setlength{\tabcolsep}{3pt}
\caption{
Average token cost of one query on HiMA-Ecom. Prompt tokens measure the context a system assembles per query, accumulated over all steps and all participating agents.
}
\begin{tabular}{lccc}
\toprule
Method & Prompt & Completion & Total \\
\midrule
GPT-4o (single agent) & 1939 & 165 & 2104 \\
DeepSeek-R1 (single agent) & 1926 & 771 & 2697 \\
DeepSeek-V3 (single agent) & 2437 & 358 & 2795 \\
Qwen2.5-14B (single agent) & 2359 & 217 & 2576 \\
Multiagent Debate (14B, 3 roles) & 7882 & 682 & 8564 \\
Qwen2.5-7B (SFT) & 9720 & 955 & 10675 \\
\textbf{HiMA-R1} (3B master + 3B subs) & 8576 & 762 & 9338 \\
\bottomrule
\end{tabular}
\label{token_cost}
\end{minipage}
\hfill
\begin{minipage}[t]{0.39\textwidth}
\centering
\setlength{\tabcolsep}{3pt}
\caption{Training time (h) under different settings of HiMA-R1 (3B), to a stable checkpoint and per checkpoint.}
\begin{tabular}{lcc}
\toprule
Setting & To stable & Per ckpt. \\
\midrule
HiMA-R1 (3B) & 9.95 & 1.11 \\
$M_{5}$ (no memory) & 16.29 & 0.81 \\
$M_{4}$ (no efficiency) & 20.61 & 1.37 \\
$M_{3}$ (update all) & 8.00 & 1.14 \\
\bottomrule
\end{tabular}
\label{training_time}
\end{minipage}
\vspace{-3mm}
\end{table}

\subsection{The cost of the accumulated context}\label{context_cost}

Table~\ref{token_cost} reports the average token consumption per query. A single agent answering the same queries under the same prompts consumes about 2.4K prompt tokens, whereas HiMA-R1 consumes 8.6K, roughly $3.6\times$ more, because tool descriptions, recalled memories, and inter-agent messages are re-assembled at every step and for every participating agent. Over 7.8 rounds per query this is about 1.1K prompt tokens per round, so the cost reflects accumulation across calls rather than a single long window.
The cost is not monotone in system complexity, since the Qwen2.5-7B SFT baseline consumes 9.7K prompt tokens while scoring 4.8 points lower, and Multiagent Debate a comparable 7.9K while scoring about 10 points lower.
Accumulated context therefore buys accuracy only when the system is trained to use it, and the components examined below decide what enters the prompt and how long a trajectory is allowed to grow.

\begin{table}[t]
\centering
\scriptsize
\setlength{\tabcolsep}{4pt}
\caption{
Ablation study on HiMA-Ecom, covering whether to train with RL rather than SFT, to generate the think process, to update top-$K$ models rather than all, to use efficiency rewards, and to integrate memory. `Update Steps' corresponds to the best-performing model.
}
\begin{tabular}{lccccccccc}
\toprule
Method & RL & Think & Top-$K$ & Efficiency & Memory & Update Steps$\downarrow$ & Accuracy$\uparrow$ & Rounds$\downarrow$ \\
\midrule
$M_{1}$ & \XSolidBrush & \XSolidBrush & - & - & -  & - & 17.2 & \textbf{5.4}\\
$M_{2}$ & \XSolidBrush & \CheckmarkBold & - & - & - & - & 35.0 & 7.1\\
\midrule
$M_{3}$ & \CheckmarkBold & \CheckmarkBold &  \XSolidBrush & \CheckmarkBold & \CheckmarkBold & 1380 & 40.0 & 7.4\\
$M_{4}$ & \CheckmarkBold & \CheckmarkBold & \CheckmarkBold & \XSolidBrush & \CheckmarkBold & 1750 & 42.4 & 8.1\\
$M_{5}$ & \CheckmarkBold & \CheckmarkBold & \CheckmarkBold & \CheckmarkBold & \XSolidBrush & 2464 & 40.0 & 7.7 \\
\midrule
$M_{6}$ & \CheckmarkBold & \CheckmarkBold & \CheckmarkBold & \CheckmarkBold & \CheckmarkBold & \textbf{1112} & \textbf{44.0} & 7.8 \\
\bottomrule
\end{tabular}
\label{contactablation}
\vspace{-3mm}
\end{table}

\subsection{Ablation study}
Table~\ref{contactablation} reports ablations based on Qwen2.5-3B agents.

\textbf{Reasoning and joint RL.}
$M_1$ trains each agent separately via SFT and is the weakest at 17.2\%, adding explicit think generation in $M_{2}$ raises it to 35.0\%, and joint RL in $M_{6}$ lifts it further to 44.0\%, obtained from the final system response alone without any agent-level supervision.

\begin{wrapfigure}{r}{0.45\textwidth}
\vspace{-3mm}
\centering
\includegraphics[width=0.45\textwidth]{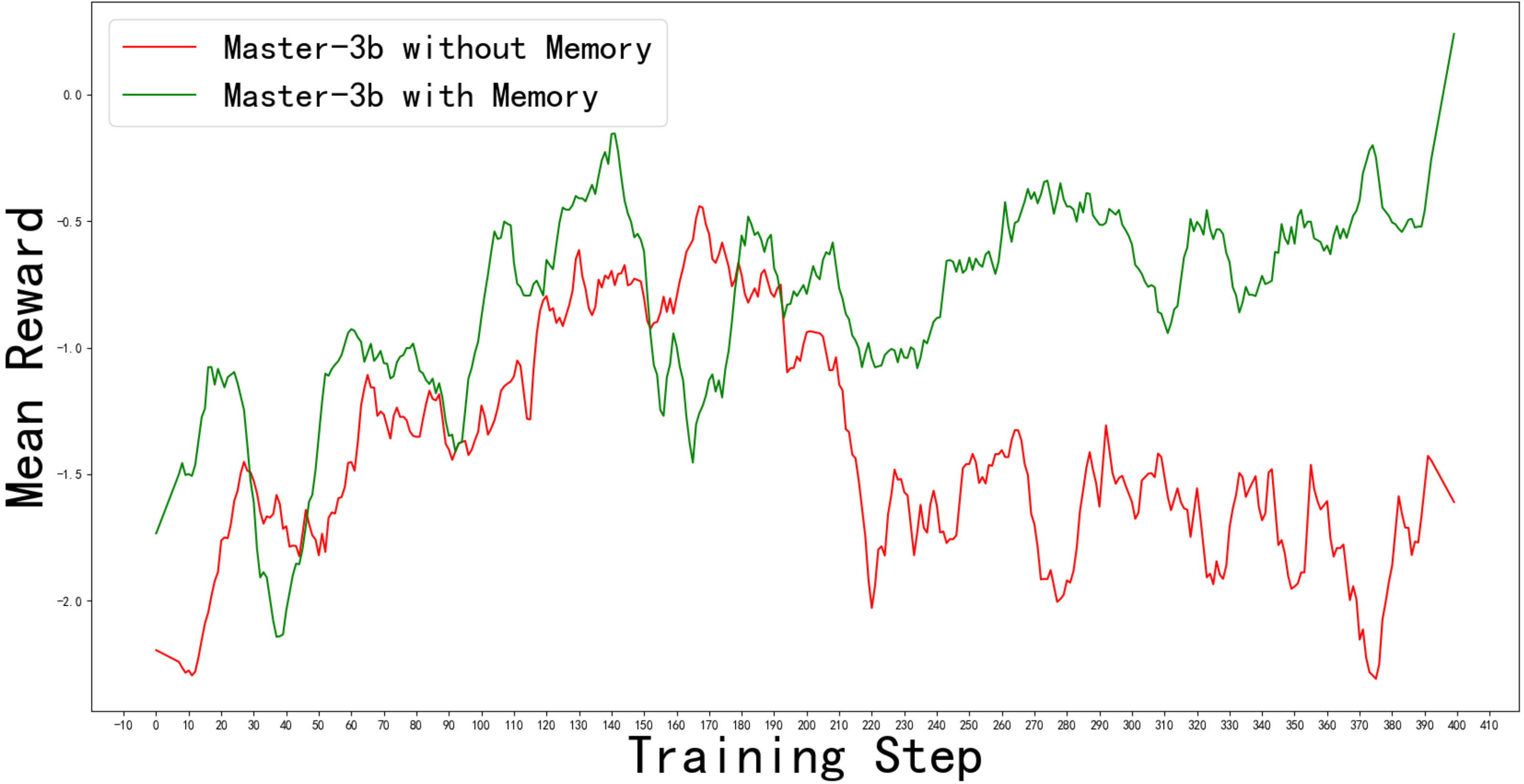}
\vspace{-5mm}
\caption{HiMA-R1 with memory achieves higher reward peaks faster during training.}
\vspace{-3mm}
\label{memory_curve}
\end{wrapfigure}
\textbf{Memory as context curation.}
$M_{5}$ disables memories across all agents. Accuracy drops from 44.0\% to 40.0\% and the update steps needed to reach the best model rise from 1112 to 2464, so memory removes roughly 55\% of the steps to convergence.
Figure~\ref{memory_curve} shows the same effect during training, where the run with memory is more stable and peaks at step 140 against step 168 without memory. In wall-clock time (Table~\ref{training_time}), each checkpoint takes longer with memory, 1.11 against 0.81 hours, since recalled entries lengthen every prompt, yet the time to a stable checkpoint drops from 16.29 to 9.95 hours, about 39\%. The number of rounds per query is nearly unchanged at 7.8 against 7.7, so the gain comes from easier optimization rather than shorter trajectories.
Curating what is recalled into the prompt therefore does not merely save prompt space, it also makes the reward signal easier to fit.

\textbf{Efficiency reward as trajectory-length control.}
$M_{4}$ removes $\mathcal{R_{E}}$ in Eq.~\ref{eq_total_reward} and shows both a lower accuracy of 42.4\% and longer trajectories with 8.1 reasoning rounds against 7.8, together with more update steps at 1750. Since every extra round appends tool responses and history to the context of every downstream agent, this term directly bounds how much context a single query can accumulate. $M_{1}$ uses the fewest rounds only because its insufficient reasoning capacity forces premature termination.

\begin{table}[t]
\centering
\scriptsize
\begin{minipage}[t]{0.5\textwidth}
\centering
\setlength{\tabcolsep}{3pt}
\caption{Ablations (\%) on updating top-$K$ models.}
\begin{tabular}{lcccccc}
\toprule
Method & Math & QA & E-FC & G-FC & Collab. & Avg. \\
\midrule
Top-$1$ & 67.0 & 16.0 & \textbf{48.0} & 75.0 & 4.0 & 42.0\\
Top-$2$ & 57.0 & 12.0 & 41.0 & 75.0 & \textbf{6.0} & 38.2\\
Top-All & 64.0 & 17.0 & 44.0 & 72.0 & 3.0 & 40.0\\
\midrule
Top-$5$ & \textbf{68.0} & \textbf{22.0} & \textbf{48.0} & \textbf{76.0} & \textbf{6.0} & \textbf{44.0} \\
\bottomrule
\end{tabular}
\label{top_k_results}
\end{minipage}
\hfill
\begin{minipage}[t]{0.48\textwidth}
\centering
\setlength{\tabcolsep}{4pt}
\caption{Ablations (\%) on number of sub-agents.}
\begin{tabular}{lcccc}
\toprule
Method & QA & E-FC & Math & Collab. \\
\midrule
Master + 2 subs & \textbf{26.0} & \textbf{55.0} & - & - \\
Master + 3 subs & 25.0 & 51.0 & \textbf{69.0} & 5.0 \\
Master + 4 subs & {22.0} & {48.0} & {68.0} & \textbf{6.0} \\
\bottomrule
\end{tabular}
\label{num_sub_results}
\end{minipage}
\vspace{-3mm}
\end{table}

\textbf{Selective updating.}
$M_{3}$ sequentially updates all models and reaches a stable checkpoint fastest in wall-clock time at 8.0 hours, but at a lower accuracy of 40.0\%, while $M_{6}$ reaches 44.0\% with the fewest update steps at 1112.
Table~\ref{top_k_results} compares different numbers of updated nodes along a trajectory, where updating the top-5 nodes performs best on average and answers 22 QA items correctly, against 16 for top-1 and 17 for updating all nodes.
Updating one or two nodes is too local to address systemic weaknesses, while updating all of them spends optimization on agents that are already stable.

\textbf{Number of sub-agents.}
Table~\ref{num_sub_results} varies how many sub-agents the master can call. Fewer sub-agents help on individual tasks, with 26 against 22 correct on QA and 55 against 48 on e-commerce function calling for the two-sub-agent setup, since a shorter list of callable sub-agents means both a shorter context for the master and less interference.
On collaborative tasks the trend reverses and the full set performs best, which is the configuration these tasks are designed for.

\section{Conclusion and Limitations}

\textbf{Conclusion.}
This paper studies hierarchical multi-agent assistants as systems whose context accumulates, since role instructions, tool documentation, recalled memories, dialogue history, and tool responses are re-assembled at every step and for every participating agent.
We introduce HiMA-Ecom, a benchmark in which 17.7K of 22.8K instances carry memory in context and which pairs agent-specific memory-augmented SFT data with system-level RL supervision, together with HiMA-R1, a recipe that keeps the resulting context budget under control through additive trajectory sampling, variance-based selective updates, an efficiency reward, and reward-driven memory evolution.
Agents built upon 3B and 7B open-source models become comparable to substantially larger LLMs, and curating what is recalled into the prompt cuts the update steps to convergence by roughly 55\%.

\textbf{Limitations.}\label{limitation}
Three directions remain open for this work.
\begin{itemize}[leftmargin=*,itemsep=1pt,topsep=2pt]
\item \textit{Evaluation granularity.} Scoring only the final answer hides where a hierarchy fails and leaves memory reuse entangled with generalization. Per-agent routing accuracy, parameter-level tool correctness, and probes that separate recalled solutions from newly derived ones would make such systems easier to compare and would locate the coordination failures behind the low collaboration scores. Our ablation also removes memory entirely, so how reward-driven evolution compares with static, similarity-based, or FIFO and LRU pools is still to be measured.
\item \textit{Scale and scope.} This study covers small to medium-scale open-source backbones in a single vertical domain, with trajectories of about eight rounds whose cost we report as prompt tokens accumulated per query rather than as the largest single context. Extending the benchmark to longer trajectories, larger memory pools, and further domains would show how far reward-driven context management carries.
\item \textit{Training infrastructure.} Heterogeneous co-training remains costly with current infrastructures, since exploring the joint action space of long trajectories is infeasible without restricting the search, and systems built for this setting would let the community study coordination at a larger scale.
\end{itemize}

{\small
\bibliographystyle{unsrtnat}
\bibliography{hima_cite}
}

\newpage
\appendix

\section{Appendix}\label{Supp}

This supplementary material is organized as follows to support the main manuscript:
\begin{itemize}[leftmargin=*]
    \item \textbf{Section~\ref{Related}:} Extended discussion of related work.
    \item \textbf{Section~\ref{Paradigm}:} Sampling strategies for hierarchical multi-agent systems.
    \item \textbf{Section~\ref{Implementation}:} More training hyperparameters and implementation details.
    \item \textbf{Section~\ref{Objective}:} The full training objective and the pipeline of VR-GRPO.
    \item \textbf{Section~\ref{Memalg}:} The dynamic memory updating algorithm.
    \item \textbf{Section~\ref{comparative_results}:} Additional comparative results on the ToolBench benchmark.
    \item \textbf{Section~\ref{Dataset}:} More details and instantiated examples of the HiMA-Ecom benchmark.
    \item \textbf{Section~\ref{Prompt}:} All prompts used by the baselines and by our multi-agent architecture.
    \item \textbf{Section~\ref{Case}:} Qualitative case analyses of agent reasoning and memory updating.
    \item \textbf{Section~\ref{LLM}:} Disclosure regarding the use of AI assistants in manuscript preparation.
\end{itemize}

\begin{table}[b]
\centering
\caption{Training settings for supervised fine-tuning.}
\begin{tabular}{c|c}
\toprule
Hyperparameter & Value \\
\midrule
learning\_rate & 5e-6 \\
max\_length & 16384 \\
num\_train\_epochs & 5 \\
max\_grad\_norm & 1 \\
weight\_decay & 0.01 \\
warmup\_ratio & 0.03 \\
lr\_scheduler\_type & cosine \\
optim & adamw\_torch \\
gradient\_accumulation\_steps & 4 \\
dataloader\_num\_workers & 8 \\
per\_device\_train\_batch\_size & 1 \\
\bottomrule
\end{tabular}
\label{hyperparams_sft}
\end{table}

\begin{table}[b]
\centering
\caption{Training settings for reinforcement learning.}
\begin{tabular}{c|c}
\toprule
Hyperparameter & Value \\
\midrule
learning\_rate & 1e-6 \\
max\_length & 16384 \\
num\_train\_epochs & 5 \\
max\_grad\_norm & 1 \\
adam\_epsilon & 1e-5 \\
num\_groups & 5 \\
topk\_groups & 5 \\
kl\_coef & 0 \\
grpo\_epoch & 2 \\
policy\_clip\_eps & 0.2 \\
temperature & 1.2 \\
per\_device\_train\_batch\_size & 1 \\
\bottomrule
\end{tabular}
\label{hyperparams_rl}
\end{table}

\subsection{Related work}\label{Related}

\textbf{LLM-based multi-agent planning.}
Recent breakthroughs in LLMs have transformed the landscape of agent planning~\cite{huang2024understanding}.
Autonomous agents implement iterative self-reflection, dynamically integrate external information via structured prompts~\cite{yao2023react, shinn2023reflexion, zhou2024language, lin2023swiftsage, qiao2024autoact}, and perceive environments to plan tasks through reasoning and decision-making~\cite{ghallab2004automated}.
Compared with single-agent approaches, multi-agent systems coordinate agents with distinct capabilities in fields like robotics~\cite{kannan2024smart}, tool calling~\cite{shen2024small}, and AI assistants~\cite{fu2024camphor}, while open-source LLMs~\cite{grattafiori2024llama, yang2024qwen2} still lag behind models that are closed-source with opaque mechanisms~\cite{achiam2023gpt, AnthropicClaude3} or overly complex for multi-agent deployment~\cite{guo2025deepseek}.
Recent work models agents as computational graphs, such as GPTSwarm~\cite{zhuge2024gptswarm}, which optimizes orchestration via prompt refinement and graph connectivity, and AFLOW~\cite{zhang2024aflow}, which automates workflow generation with Monte Carlo Tree Search, whereas modular frameworks such as OxyGent~\cite{hu2026oxygent} assemble agents and tools into execution graphs at runtime, while Multiagent Debate~\cite{du2024improving} prompts multiple LLMs to iterate a debate, and such coordination also brings new failure modes across perception, communication, and reasoning~\cite{zhou2026hierarchical}.
This work studies a hierarchical architecture that interprets user queries and performs dynamic planning under the context accumulated in a vertical domain.

\textbf{Multi-agent reinforcement learning.}
MARL has witnessed substantial advancements for complex tasks~\cite{yuan2025efficient}, and this work focuses on cooperative settings where agents share a common goal, as applied in game playing~\cite{wang2021rode, yu2022surprising}, task allocation~\cite{iqbal2022alma}, skill discovery~\cite{yang2023hierarchical}, and circuit design~\cite{wang2024hierarchical}.
Typical MARL methods employ an actor-critic framework~\cite{sun2024llm}, including policy-based methods like MADDPG~\cite{lowe2017multi} and MAPPO~\cite{yu2022surprising} and value-based ones like VDN~\cite{sunehag2018value} and QMIX~\cite{rashid2020monotonic}.
LLM-based MARL frameworks for problem-solving~\cite{chen2023multi, hong2024metagpt} and embodied intelligence~\cite{mandi2024roco, zhang2024building, kannan2024smart} mainly enhance inter-agent communication rather than joint evolution, and parameter sharing across agents~\cite{deng2024smac, terry2020revisiting, christianos2021scaling} restricts applicability to homogeneous scenarios~\cite{zhong2024heterogeneous}.
GRPO~\cite{shao2024deepseekmath} eliminates the value function and relies on observed rewards, which suits heterogeneous agents~\cite{ke2025survey}, and MLPO~\cite{estornell2025train} trains the leader of a multi-agent system, while ReMA~\cite{wan2025rema} and MARFT~\cite{liao2025marft} train agents with multi-agent RL on reasoning-style tasks. Such training is also frequently combined with parameter-efficient adaptation~\cite{hu2021lora, zhang2026primacy, zhang2026hidden, wang2026tina}.
The setting studied here differs in that agents are functionally distinct, each carries its own memory-augmented context, and only the final system response is supervised, so sampling and update selection are used as practical devices for keeping joint training affordable.

\textbf{LLM-based agent memory.}
Agent memory can be divided into RAG-based and embodied categories~\cite{huang2024understanding}, where the former is kept in additional storage and the latter embeds memories into model parameters.
This work focuses on RAG-based long-term memory, in which recalled entries are placed directly into the prompt and therefore compete for the context budget of every step.
Recent works have explored diverse strategies~\cite{packer2023memgpt, lu2023memochat, wang2023augmenting, huang2023memory}, where MoT~\cite{li2023mot}, TiM~\cite{liu2023think}, and RAP~\cite{kagaya2024rap} leverage memories after selection or thinking, MemoryBank~\cite{zhong2024memorybank} draws on the Ebbinghaus forgetting curve~\cite{ebbinghaus2013memory} for selective retention, and HELPER~\cite{sarch2023open}, ExpeL~\cite{zhao2024expel}, RET-LLM~\cite{modarressi2024ret}, Synapse~\cite{zheng2024synapse}, and A-mem~\cite{xu2025mem} differ in knowledge aggregation, storage, and retrieval.
Memory mechanisms tailored for multi-agent systems~\cite{zhang2024survey} further study synchronization~\cite{chen2024scalable}, communication~\cite{mandi2024roco}, and information asymmetry~\cite{light2023text} among agents.
In contrast, we let memory and decision modules evolve synergistically with LLM optimization, so that what stays in context is decided by the same signal that trains the policies.

\subsection{Sampling strategies}\label{Paradigm}

\begin{figure}[h]
\centering
\includegraphics[width=0.9\textwidth]{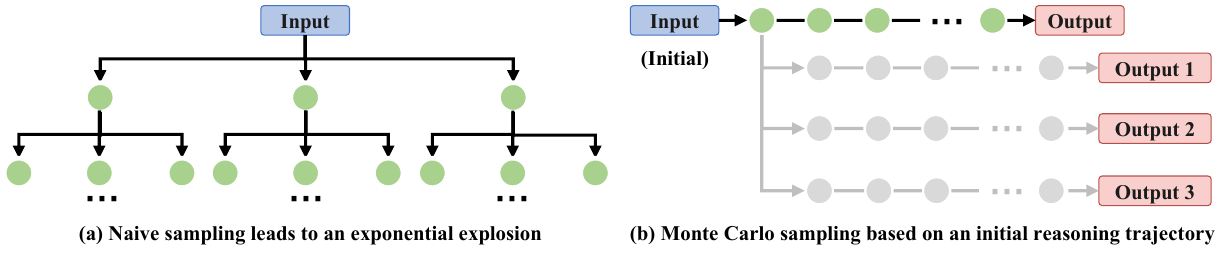}
\caption{
Different sampling strategies for multi-agents.
Green circles indicate nodes sampling multiple actions via GRPO, while gray circles involve no additional sampling.
Compared to naive sampling in (a), (b) could avoid the exponential explosion of the multi-agent joint action space.
}
\label{sampling}
\end{figure}

\subsection{Implementation details}\label{Implementation}

All agents are fully fine-tuned, so every parameter is updated during both the SFT and the RL stages, without adapters, low-rank modules, or parameter sharing across agents. At each step, the top-5 memories retrieved for the current query are placed in the context of the acting agent.
In this work, each agent from an initial trajectory is sampled $G_i = 5$ actions with a temperature of $1.2$. Subsequently, the top-$5$ nodes are selected for model updates. Similar to DeepSeek-R1~\cite{guo2025deepseek}, the iterative RL with GRPO is executed for 2 iterations. Regarding memory evolution, the deletion threshold is set to $D = 0$, with hyperparameters $\alpha = \beta = 1$.
To improve the reproducibility of the experiments, more training settings for supervised fine-tuning (SFT) and reinforcement learning (RL) are provided in Table~\ref{hyperparams_sft} and Table~\ref{hyperparams_rl}, respectively.
To compare the SFT and RL methods fairly, the former is trained for a total of 10 epochs, while the latter is trained with SFT for 5 epochs followed by RL for 5 epochs.
For RL training, the models are deployed on the TRL~\cite{vonwerra2022trl} framework on 2 NVIDIA H200 GPUs for accelerated inference via vLLM~\cite{kwon2023efficient} and real-time weight updates, while allocating 6 GPUs for joint training.

\subsection{Training objective and pipeline}\label{Objective}

Given a query-answer pair $(q,a)$, the old policy $\pi_{\theta_i\text{old}}$ of each agent $m_i$ samples a group of outputs $\{ o_j\}_{j=1}^{G_i}$, and the policy model $\pi_{\theta_i}$ is optimized by maximizing
\begin{equation}
\begin{split}
    & \mathcal{J}_{GRPO}(\theta_i) = \mathbb{E}{[q \sim P(Q), \{o_j\}_{j=1}^{G_i} \sim \pi_{\theta_i{old}}(O|q)]}  \\
    & \frac{1}{G_i}\sum_{j=1}^{G_i} \left( \min \left( \frac{\pi_{\theta_i}(o_j |q)}{\pi_{\theta_i{old}}(o_j |q)} A_j, \text{clip} \left( \frac{\pi_{\theta_i}(o_j |q)}{\pi_{\theta_i{old}}(o_j |q)}, 1 - \epsilon, 1 + \epsilon \right)  A_j \right)\right),
\end{split}
\label{eq_GRPO_single}
\end{equation}
where $\epsilon$ is a hyper-parameter for the clipped objective and $A_j$ denotes the advantage of the $j$-th response, obtained by normalizing the group-wise rewards with their average and standard deviation. Updating every agent of the trajectory corresponds to $\mathcal{J}_{Multi\text{-}GRPO}(\Theta) = \{\mathcal{J}_{GRPO}(\theta_i) \mid i = 1, \cdots ,k \}$, whereas VR-GRPO restricts the update as in Eq.~\ref{eq_GRPO_V}.

\subsection{Memory updating algorithm}\label{Memalg}

Algorithm~\ref{alg_memory} details the dynamic memory updating procedure described in Section~\ref{grpomemory}. Memories recalled by the current query are rescored according to their similarity to the trajectory output and the trajectory reward, while all other memories decay with time.

\begin{algorithm}[h]
\small
\caption{Dynamic Memory Updating from GRPO Rewards}
\label{alg_memory}
\textbf{Input}: Planning chain $\mathcal{P}$, Trajectory output $\mathcal{O}$, Trajectory reward $\mathcal{R_{M}}$, Upper bound $U$, Lower bound $L$, Set of $n$ recalled memories $Recall_q$ given query $q$, Hyperparameters $\alpha$ and $\beta$, Timestamp $t$
\begin{algorithmic}[1]
\For{each memory $\mathcal{M}_i \in Recall_q$}
    \State Compute similarity: ${sim}_i \leftarrow {Sim}(\mathcal{O}, \mathcal{O}_i)$ (Direct answer mode) \textbf{or} ${Sim}(\mathcal{P}, \mathcal{P}_i)$ (Tool call mode)
    \If{$\mathcal{R_{M}} > U$}
        \State Update timestamp: $t_i \leftarrow t$;
        \State Compute time and reward differences: $\Delta t \leftarrow -|t - t_i|$ \textbf{and} $\Delta s \leftarrow s_i \cdot |\mathcal{R_{M}} - U|$
    \ElsIf{$\mathcal{R_{M}} < L$}
        \State Compute time and reward differences: $\Delta t \leftarrow -|t - t_i|$ \textbf{and} $\Delta s \leftarrow - s_i \cdot |\mathcal{R_{M}} - L|$;
        \State Update timestamp: $t_i \leftarrow t$
    \EndIf
    \State Update recalled memory scores: $\mathcal{R_M}_{i} \leftarrow \mathcal{R_M}_{i} + \alpha \Delta t + \beta \Delta s$
\EndFor
\For{each memory $\mathcal{M}_i \notin Recall_q$}
    \State Compute time difference: $\Delta t \leftarrow -|t - t_i|$;
    \State Update timestamp: $t_i \leftarrow t$;
    \State Update recalled memory scores: $\mathcal{R_M}_{i} \leftarrow \mathcal{R_M}_{i} + \alpha \Delta t$
\EndFor
\end{algorithmic}
\end{algorithm}

\subsection{Additional comparative results}\label{comparative_results}

\paragraph{Experimental setup for the ToolBench dataset.}
More comparative experiments are conducted on ToolBench~\cite{qin2024toolllm}, which involves integrating API calls to accomplish tasks, where the agent should select the correct API and compose the necessary API requests accurately.
The test set is divided into in-domain and out-of-domain parts based on whether the tools used in the test instances have been seen during training, which evaluates both learning and generalization.
The proposed method is compared with two supervised baselines built on Qwen2.5-3B and fine-tuned on the ToolBench training set, namely HiMA-SFT, which outputs the reasoning process and the final result, and HiMA-SFT-no, which directly generates the final response without a thinking process. Larger open- and closed-source LLMs~\cite{yang2024qwen2, guo2025deepseek, deepseekai2024deepseekv3, achiam2023gpt} are also used to build agents for comparison without fine-tuning.
The metrics include Plan\_ACC, the accuracy of planning decisions at each step of tool calling; Act\_EM, the proportion of predicted API names that exactly match the real ones; Easy\_F1 and Hard\_F1, the argument F1 scores when the ground truth argument is empty and non-empty respectively; F1, the argument F1 across all conditions; No\_Hallu, the frequency of predicted API names without hallucination; and their average.

\begin{table}[h]
  \centering
  \scriptsize
  \setlength{\tabcolsep}{4pt}
\caption{Comparative results (\%) on the ToolBench dataset.
Bold represents the optimal score, and `\_' represents the suboptimal value. `HiMA-SFT-no' means training without a thinking process.
}
  \begin{tabular}{lccccccc}
  \toprule
    Method & Plan\_ACC  & Act\_EM  & Easy\_F1 & Hard\_F1 & F1 & No\_Hallu & Average  \\
    \midrule
    \multicolumn{8}{c}{\textbf{In the domain}}\\
    \midrule
    GPT-4o & \textbf{82.5} & \textbf{57.0} & \textbf{31.1} & 22.2 & \underline{37.3} & \underline{99.9} & \textbf{55.0}  \\
    DeepSeek-R1 & 72.3 & 45.2 & \underline{18.7} & 18.2 & 28.5 & \textbf{100.0} & 47.1  \\
    DeepSeek-V3 & 68.4 & 38.0 & 14.9 & 14.3 & 23.6 & 99.4 & 43.1  \\
    Qwen2.5-32B & 72.7 & 43.7 & 15.1 & 21.0 & 28.8 & 99.3 & 46.8  \\
    Qwen2.5-14B & 42.2 & 15.0 & 3.7 & 7.1 & 10.1 & \textbf{100.0} & 29.7  \\
    HiMA-SFT-no (7B+3B) & 55.7 & 33.0 & 10.1 & 20.5 & 26.5 & 95.7 & 40.3  \\
    HiMA-SFT (7B+3B) & 73.3 & 47.0 & 16.3 & \underline{29.6} & 36.8 & 94.5 & 49.6  \\
    HiMA-R1 (3B+3B) & \underline{73.5} & \underline{48.5} & 17.6 & \textbf{31.5} & \textbf{39.0} & 96.7 & \underline{51.2} \\
    \midrule
    \multicolumn{8}{c}{\textbf{Out of the domain}}\\
    \midrule
    GPT-4o & \textbf{83.9} & \textbf{63.2} & \textbf{42.9} & 22.3 & 44.1 & 99.8 & \textbf{59.3}  \\
    DeepSeek-R1 & 74.6 & 52.0 & 28.3 & 20.4 & 35.6 & 99.8 & 51.8  \\
    DeepSeek-V3 & \underline{77.2} & 56.1 & 32.0 & 21.9 & 38.2 & \textbf{100.0} & 54.2  \\
    Qwen2.5-32B & 28.6 & 0.0 & 0.0 & 0.0 & 0.0 & \textbf{100.0} & 21.4  \\
    Qwen2.5-14B  & 43.9 & 16.6 & 7.2 & 6.2 & 12.3 & \underline{99.9} & 31.0  \\
    HiMA-SFT-no (7B+3B) & 59.7 & 41.1 & 19.7 & 25.9 & 36.5 & 97.3 & 46.7  \\
    HiMA-SFT (7B+3B) & 72.3 & 57.0 & \underline{32.3} & \underline{36.1} & \underline{48.7} & 96.3 & 57.1  \\
    HiMA-R1 (3B+3B) & 73.4 & \underline{57.5} & 31.1 & \textbf{37.0} & \textbf{48.9} & 96.9 & \underline{57.5} \\
  \bottomrule
  \end{tabular}
  \label{toolbench_results}
\end{table}

\paragraph{Comparative results on the ToolBench dataset.}
Table~\ref{toolbench_results} shows consistent results across in-domain and out-of-domain settings.
In-domain, HiMA-R1 outperforms both DeepSeek-R1 and DeepSeek-V3 in Plan\_ACC, Act\_EM, Hard\_F1, F1, and average score, reaching 51.2\% against 47.1\% and 43.1\% with only 3B agents, and it surpasses GPT-4o on Hard\_F1 and F1 both in-domain (31.5\% and 39.0\%) and out-of-domain (37.0\% and 48.9\%).
Enabling the think process improves the SFT baseline from 40.3\% to 49.6\%, with a further 3.2\% relative gain through multi-agent reinforcement learning.

\subsection{More details of the HiMA-Ecom benchmark}\label{Dataset}

The proposed HiMA-Ecom does not contain any personally identifiable or sensitive private information.
The benchmark can be divided into two categories, SFT and RL. Since the RL stage is trained end-to-end, this type of data only contains the initial query and the final response. More details of the SFT data used for each agent are introduced as follows, where Tables~\ref{table_case_1} to~\ref{table_case_3} are SFT training instances, so the memories recalled in them belong to the training data:

\textbf{Master agent.}
As shown in Table~\ref{table_case_1}, the case for the master agent includes user queries, optional tools, invoked agents, retrieved memories, historical dialogues, and tool-generated responses.
In addition, the reasoning processes (\emph{i.e.}, $<\boldsymbol{think}> \cdots <\boldsymbol{/think}>$ ), tool calling (\emph{i.e.}, $<\boldsymbol{tool\_call}> \cdots <\boldsymbol{/tool\_call}>$ ), and final answers are provided in the benchmark.

\textbf{Function-call agents.}
Regarding function-calling, the e-commerce agent incorporates 12 prevalent APIs of e-commerce platforms, while the general agent draws from 3.3K diverse function calls from ToolBench~\cite{qin2024toolllm}.
Table~\ref{table_case_2} illustrates a case for the e-commerce function-call agent, which appears as `rody\_agent' in the prompts and code.

\textbf{Question-answering agent.}
The question-answering agent includes `tool\_retrievals' and `tool\_summary' as demonstrated in Table~\ref{table_case_3}. Its data covers open-domain (COIG~\cite{zhang2023chinese}) and e-commerce scenarios.
The QA agent appears as `expert\_agent' in the prompts and code.

\textbf{Math agent.}
The dataset for the math agent is constructed from GSM8K~\cite{cobbe2021training}. The last part of Table~\ref{table_case_1} shows the thinking process and answers of math agents.

\onecolumn
\renewcommand{\arraystretch}{1.3}
\begin{longtable}{m{2cm}<{\centering}| m{11cm}}
\caption{A case of the master agent in a collaborative task.}\label{table_case_1}\\
\toprule
\multicolumn{2}{c}{\textbf{
Case \#1 of the master agent in a collaborative task
}} \\ \hline
\addlinespace
\multicolumn{2}{p{370pt}}{\textbf{User:} I want to open a small personal store on [e-commerce platform A] to sell educational toys. How much deposit do I need to pay? If I open two stores, how much deposit will I need to pay in total?}\\
\addlinespace
\hline

\textbf{System}  &
\textbf{\{Role\}} from Template \#1 in Section~\ref{Prompt}

\textbf{Specialized Retrieval Tools/Memory:}

$<\boldsymbol{tools}> $

[\{"name": "math\_agent",
"description": "When the user puts forward a clear request to give a solution or answer to a specific mathematical problem, the agent needs to be called. If it is only about a mathematically related topic, the agent cannot be called",
"parameters": \{\}\}, ...]

$<\boldsymbol{/tools}> $

$<\boldsymbol{memory}> $

[\{"ID": 1620,
"Query": " Calculate how many times Danny folded his picnic blankets given the initial and final area.",
"Plan": ["math\_agent"],
"Answer": "How many square feet are each blanket? ** The total area of each blanket is 8*8=<<8*8=64>>64.
How many square feet are the blankets? ** Their total area is 192 square feet because 3 x 64 = <<3*64=192>>192. How many times did he fold them? ** He folded them 4 times because 192 / 48 = <<192/48=4>>4. \#\#\#\# 4",
"Time": "2025-03-26 11:41:13",
"Score": 6\}
...]

$<\boldsymbol{/memory}> $

\textbf{\{Problem Resolution Framework\}} from Template \#1 in Section~\ref{Prompt}\\
\hline

\textbf{Assistant}  &
$<\boldsymbol{tool\_call}> $

\{"name": "rody\_agent",
"arguments": \{"intention": ["Query the deposit required for opening a small personal store on [e-commerce platform A] to sell educational toys."]\}\}

$<\boldsymbol{/tool\_call}> $
\\ \hline

\textbf{Tool}  &
Pass calling
\{"api\_name": "check\_shop\_expenses",
"parameter": \{"keyword": ["Educational Toys"], "shop\_body": ["Individual"], "shop\_type": ["small shops"]\}\}, the API call was successful.

Results are as follows:

1. Specific amount of margin per year: 6000 yuan.

2. Transaction service fee rate per transaction: 0.25\%.

3. Technical service fee rate per transaction: 0.41\%.
\\ \hline

\textbf{Assistant}  &
$<\boldsymbol{tool\_call}> $

\{"name": "tool\_retrievals",
"arguments": \{"intention": ["Calculate the total deposit required for opening two small personal stores on [e-commerce platform A] to sell educational toys, based on the queried deposit information of 6000 yuan per store."]\}\}

$<\boldsymbol{/tool\_call}> $
\\ \hline

\textbf{Tool}  &
[\{"name": "expert\_agent",
"description": "When users chat, greet, abuse, make jokes, discuss music/games/current politics/military/entertainment/food and other daily topics, or consult the rules of the e-commerce platform, the agent needs to be invoked ",
"parameters": \{\}\},

\{"name": "math\_agent",
"description": "When the user puts forward a clear request to give a solution or answer to a specific mathematical problem, the agent needs to be called. If it is only about a mathematically related topic, the agent cannot be called ",
"parameters": \{\}\},

\{"name": "toolbench\_agent",
"description": "function-0: This is the subfunction for tool "football\_xg\_statistics" and its function is "seasons list by tournamentid for football xg statistics", you can use this tool. The description of this function is: Football (soccer) xG statistics. We provide statistics for more than 80 leagues. There are more than 90,000 games and 2 million shots in our database. The method is used to get a list of seasons by the league ID; function-1: This is the subfunction for the tool "mailslurp\_email\_testing" and its function is "getbouncedrecipients for mailslurp email testing", you can use this tool. The description of this function is: Email sandbox testing API. Bounced recipients are email addresses that you have sent emails to that did not accept the email. Once a recipient is bounced, you cannot send emails to that address."\}]
\\ \hline

\textbf{Assistant}  &
$<\boldsymbol{tool\_call}> $

\{"name": "math\_agent",
"arguments": \{"intention": ["Calculate the total deposit required for opening two small personal stores on [e-commerce platform A] to sell educational toys, based on the queried deposit information of 6000 yuan per store."]\}\}

$<\boldsymbol{/tool\_call}> $

\\
\hline

\textbf{Tool}  &

The user wants to open a small personal store on [e-commerce platform A] to sell educational toys. The deposit required for one store is 6000 yuan per year. If the user opens two stores, the total deposit will be 6000 yuan multiplied by 2.

12000

\\ \hline

\textbf{Assistant Final}  &
$<\boldsymbol{think}> $

The user's issue is to calculate the total deposit required for opening two small personal stores on [e-commerce platform A] to sell educational toys. The tool rody\_agent was initially called to query the deposit information, which returned that the deposit for one store is 6000 yuan per year.
The user then asked for the total deposit for two stores. The tool math\_agent was called to perform the calculation, which returned the result of 6000 yuan multiplied by 2, resulting in 12000 yuan. Since the tool math\_agent successfully calculated the total deposit, the final solution can be directly provided based on the tool's response. Therefore, the final solution is "12000".

$<\boldsymbol{/think}> $

12000\\
\hline
\end{longtable}

\onecolumn
\renewcommand{\arraystretch}{1.5}
\begin{longtable}{m{2cm}<{\centering}| m{11cm}}
\caption{A case of the function-call agent in an e-commerce task.}\label{table_case_2}\\
\toprule
\multicolumn{2}{c}{\textbf{
Case \#2 of the function-call agent in an e-commerce function-call task
}} \\ \hline
\addlinespace
\multicolumn{2}{p{370pt}}{\textbf{User:} Hello, I found that some orders may have the label of door-to-door replacement, what information should be provided in this case?}\\
\addlinespace
\hline

\textbf{System}  &
\textbf{\{Role\}} from Template \#2 in Section~\ref{Prompt}

\textbf{Specialized Retrieval Tools/Memory:}

$<\boldsymbol{tools}>$

[\{"name": "search\_payment\_method","description": "Inquire the settlement method of the user's shop, such as B2B settlement, C2B settlement, bank card settlement, online banking wallet settlement.","parameters": \{\}\},
\{"name": "search\_order\_code", "description": "...", ,"parameters": \{...\}\},
...]

$<\boldsymbol{/tools}> $

$<\boldsymbol{memory}> $

[\{"ID": "1254", "Query": "Hello, I found that some orders may have the label of door-to-door replacement, what information should be provided in this case?", "Plan": [], "Answer": "Pass calling \{'api\_name': 'search\_order\_code', 'parameter': \{'order\_service\_mix\_tag': ['door-to-door replacement'], 'search\_keys': ['order label']\}\}, the API call was successful.

Results are as follows:

We have retrieved the relevant order information for you.", "Time": "2025-04-29 02:20:30", "Score": 10\},
...]

$<\boldsymbol{/memory}> $

\textbf{\{Problem Resolution Framework\}} from Template \#2 in Section~\ref{Prompt}\\
\hline

\textbf{Assistant}  &
$<\boldsymbol{tool\_call}> $

\{"name":"search\_order\_code", "arguments":\{"order\_service\_mix\_tag": ["door-to-door replacement"], "search\_keys": ["order label"]\}\}

$<\boldsymbol{/tool\_call}> $
\\ \hline

\textbf{Tool}  &
Pass calling \{"api\_name": "search\_order\_code", "parameter":\{"order\_service\_mix\_tag": ["door-to-door replacement"], "search\_keys": ["order label"]\}\}, the API call was successful.

Results are as follows:

We have retrieved the relevant order information for you.
\\ \hline

\textbf{Assistant Final}  &
$<\boldsymbol{think}> $

The Input issue is about retrieving information for orders with the label "door-to-door replacement." The Output is the result of calling the "search\_order\_code" tool with the parameters `\{"order\_service\_mix\_tag": ["door-to-door replacement"], "search\_keys": ["order label"]\}'. The Output indicates that the API call was successful and relevant order information has been retrieved. This matches the reference information in <memory>, where a similar query was resolved by calling the same tool with the same parameters. Therefore, the Output is directly derived from the reference information, and the reasoning process is consistent with the provided solution.

$<\boldsymbol{/think}> $

Pass calling \{"api\_name": "search\_order\_code", "parameter": \{"order\_service\_mix\_tag": ["door-to-door replacement"], "search\_keys": ["order label"]\}\}, the API call was successful.

Results are as follows: We have retrieved the relevant order information for you.\\
\hline
\end{longtable}

\onecolumn
\renewcommand{\arraystretch}{1.5}
\begin{longtable}{m{2cm}<{\centering}| m{11cm}}
\caption{A case of the question-answering agent in a QA task.}\label{table_case_3}\\
\toprule
\multicolumn{2}{c}{\textbf{
Case \#3 of the question-answering agent in a QA task
}} \\ \hline
\addlinespace
\multicolumn{2}{p{370pt}}{\textbf{User:} Where to open the LCL service?}\\
\addlinespace
\hline

\textbf{System}  &
\textbf{\{Role\}} from Template \#4 in Section~\ref{Prompt}

\textbf{Specialized Retrieval Tools/Memory:}

$<\boldsymbol{tools}> $

[\{"name": "tool\_retrievals", "description": "Knowledge retrieval tool. Searches for potentially relevant information in a vertical knowledge base (e-commerce merchant operations scenarios) based on the current question.", "arguments": \{\}\}
\{"name": "tool\_summary", "description": "Intelligent Q\&A tool. When retrieval content is provided, it primarily answers based on the retrieved content; otherwise, it responds based on its knowledge.", "arguments": \{\}\}]

$<\boldsymbol{/tools}> $

$<\boldsymbol{memory}> $

[\{"id": 4439, "model\_name": "expert", "query": "Is my shop bounded with a card?", "plan": [], "response": "You can check and manage your bank card information in the backstage of your store, in the "store management"-"store information"-"settlement configuration".", "score": 1.0, "time": 7514\},
...]

$<\boldsymbol{/memory}> $

\textbf{\{Problem Resolution Framework\}} from Template \#4 in Section~\ref{Prompt}\\
\hline

\textbf{Assistant}  &
$<\boldsymbol{tool\_call}> $

\{"name": "tool\_retrievals", "arguments": \{\}\}

$<\boldsymbol{/tool\_call}> $
\\ \hline

\textbf{Tool}  &

[General] Corporation\_A Merchant Consolidation Warehouse

Opening Configuration Page (see link: ...)

[Other] Merchant Question: Consolidation Opening Entry

Merchant Assistant:

1) How to open: You can click on the left menu bar of the Merchant Center - Services - Merchant Service Center - Sales - Remote Area Logistics Service - Click to open to enter the product configuration page. Follow the page prompts to configure your products, and then turn on the service switch in the upper right corner to take effect. Direct link: ...

2) Fees: For products bound to consolidation and free shipping, you only need to bear the cost of the first leg to Xi'an. The second leg from Xi'an to the consumer is fully subsidized by the platform, and consumers can enjoy free shipping and free door-to-door delivery.

3) Benefits: Free second-leg logistics for remote area shipments, free interception for second-leg logistics in remote areas, estimated doubling of remote area order volume, and exemption from store comprehensive experience scores.
...

\\ \hline

\textbf{Assistant}  &
$<\boldsymbol{tool\_call}> $

\{"name": "tool\_summary", "arguments": \{\}\}

$<\boldsymbol{/tool\_call}> $
\\ \hline

\textbf{Tool}  &
The entry to open the consolidation service is as follows:

1. Click the left menu bar.

2. Select "Service".

3. Enter "Merchant Service Center".

4. Select "Remote Area Logistics Service" under "Selling Section".

5. Click "Open" to enter the product configuration page.

6. Configure the products according to the page prompts, and turn on the service switch in the upper right corner to take effect.

Direct link: ...
\\ \hline

\textbf{Assistant Final}  &
$<\boldsymbol{think}> $

Given that the user's question pertains to e-commerce-specific knowledge and both tool\_retrievals and tool\_summary have already been invoked in the historical context, the optimal path is to directly output the answer provided by tool\_summary. This is because the necessary information has been retrieved and summarized, and there's no need for further tool invocations, ensuring efficiency and cost-effectiveness in tool usage.

$<\boldsymbol{/think}> $

The entry to open the consolidation service is as follows:

1. Click the left menu bar.

2. Select "Service".

3. Enter "Merchant Service Center".

4. Select "Remote Area Logistics Service" under "Selling Section".

5. Click "Open" to enter the product configuration page.

6. Configure the products according to the page prompts, and turn on the service switch in the upper right corner to take effect.

Direct link: ...\\
\hline
\end{longtable}

\subsection{Prompts}\label{Prompt}

\paragraph{System prompts for multi-agents.}
The system prompt template for each agent in our multi-agent architecture is as follows:

\begin{tcolorbox}[colframe=black, colback=gray!10, title=Template \#1: System Prompt for the Master Agent, breakable]

~

\textbf{Role:}

~

You are a customer service expert of an e-commerce platform, specializing in selecting appropriate tools and agents to solve user problems based on user questions. Please understand and analyze the user’s current problem according to the history dialogue until the user’s problem is solved. There are some tools available between $<\boldsymbol{tools}>$$<\boldsymbol{/tools}>$for selection at each step. You can think as follows:

\begin{itemize}
  \item If the user’s intention can completely match other known specific tools except for tool\_intention and tool\_retrievals, directly call that tool.
  \item If you cannot understand the user’s current intention, call tool\_intention.
  \item If the user’s intention is clear but cannot match other known specific tools except for tool\_intention and tool\_retrievals, call tool\_retrievals to get complementary information to choose from all tools again.
  \item If an agent tool has been called and returned results, or content has been recalled between $<\boldsymbol{memory}>$$<\boldsymbol{/memory}>$, you need to judge whether the current agent result or the similar answers recorded in the memory meet the user’s intention. If so, directly output the answer based on it; otherwise, re-arrange.
\end{itemize}

~

\textbf{Specialized Retrieval Tools / Memory:}

~

$<\boldsymbol{tools}> $

\begin{itemize}

\item \{"name": "rody\_agent", "description": "The agent has the following functions:

\begin{itemize}
  \item Check fee: check the deposit, quality guarantee, and technical service fee requirements required for settling in the Corporation\_A platform;
  \item Check qualifications: inquire about the various documents/material requirements required for settling in the Corporation\_A platform;
  \item Check the order: Check all questions about the related order;
  \item Query after-sales information: Check all questions about the related after-sales order;
  \item Order reporting: Delayed reporting of orders that cannot be delivered on time, but unable to check whether the order has been reported or whether it needs to be reported;
  \item Check the refund of deposit at the time of check-out: When the user applies for check-out and returns the deposit, check the current refund progress of the deposit and the reason why the deposit cannot be returned at present;
  \item Query product promotion: Quickly obtain detailed information related to the designated promotion activities of the merchants;
  \item Query product coupons: obtain the status of coupons specified by the merchant and applicable products, but can not query the reason why the coupons are not effective;
  \item Check the status of bank card: Help users check the current binding and verification status of their bank card;
  \item Query the settlement method: According to the actual situation of the user, help query the settlement method of its store, and provide the current store payment and platform refund flow;
  \item Query product audit: According to the actual requirements of the user, query the review status of the user's product listing/modification and the reasons for the slow review progress, but can not query the reasons for the failure of the product audit."\}
\end{itemize}

\item \{"name": "expert\_agent", "description": "When users chat, greet, abuse, make jokes, discuss music/games/current politics/entertainment/food and other daily topics, or consult the rules of the e-commerce platform, the agent needs to be invoked.", "parameters": \{\}\}

\item \{"name": "math\_agent", "description": "When the user puts forward a clear request to give a answer to a specific mathematical problem, the agent needs to be called. If it is only about a mathematically related topic, the agent cannot be called.", "parameters": \{\}\}
\item \{"name": "toolbench\_agent", "description": "", "parameters": \{\}\}
\item \{"name": "tool\_intention", "description": "The intention understanding tool. Understand the user’s real intention based on context and current question.", "arguments": \{\}\}
\item \{"name": "tool\_retrievals", "description": "The API retrieval tool. Retrieve related APIs from the API knowledge base based on intention.", "arguments": \{"intention": "user's current intention"\}\}
\item ...

\end{itemize}

$<\boldsymbol{/tools}> $

$<\boldsymbol{memory}> $

{memory\_append}(Optional)

$<\boldsymbol{/memory}> $

~

\textbf{Problem Resolution Framework:}

~

Specific requirements are as follows:

\begin{itemize}
  \item When selecting tools, please refer to the tool’s function description. Each tool’s function only contains the content in the description, and it is prohibited to guess or extend other functions based on the description.
  \item When selecting tools after calling tool\_retrievals, you can only choose from the candidate tool set retrieved through tool\_retrievals. Each tool’s function only contains the content in the description.
  \item Please be faithful to the semantics of the current problem and historical dialogue, and do not output content that does not exist in the historical dialogue and current problem.
  \item Directly output the agent’s response if there is no error.
  \item Do not alter or truncate any words if the response is from rody/math/toolbench.
  \item For math or coding questions, use the user’s query as the sub-agent calling intention.
  \item Do not call the same tool repeatedly. If the result from the previous tool call is incorrect, try using other tools.
  \item When a user requests a solution or answer to a specific math problem, output in the following format (output the numerical answer directly after $<\boldsymbol{/think}> $, without any units or irrelevant characters): $<\boldsymbol{think}> $think process$<\boldsymbol{/think}> $answer.
  \item Output strictly according to the following format:
  \begin{itemize}
    \item The user’s question is not clear, or unable to understand the user’s intention:
    \begin{itemize}
      \item $<\boldsymbol{think}> $
      The analysis and thinking process of the user’s problem, and the reason for calling the intention recognition tool.
      $<\boldsymbol{/think}> $
      \item $<\boldsymbol{tool\_call}> $
      \{"name": "tool\_intention", "arguments": \{\}\}
      $<\boldsymbol{/tool\_call}> $
    \end{itemize}

    \item The tools in <tools></tools> and all the currently retrieved tools cannot meet the user’s intention:
    \begin{itemize}
      \item $<\boldsymbol{think}> $
      The analysis and thinking process of the user’s problem, and the reason for calling the tool\_retrievals tool.
      $<\boldsymbol{/think}> $
      \item $<\boldsymbol{tool\_call}> $
      \{"name": "tool\_retrievals", "arguments": \{"intention": user\_intention\}\}
      $<\boldsymbol{/tool\_call}> $
    \end{itemize}

    \item The agent tools in <tools></tools> or all the currently retrieved tools can meet the user’s intention:
    \begin{itemize}
      \item $<\boldsymbol{think}> $
      The analysis and thinking process of the user’s problem, and the reason for choosing which agent to solve the user's problem.
      $<\boldsymbol{/think}>$
      \item $<\boldsymbol{tool\_call}> $
      \{"name": chosen agent name, "arguments": \{"intention": user's intention that can be met through chosen agent\}\}$<\boldsymbol{/tool\_call}> $
    \end{itemize}

    \item According to the recalled information between $<\boldsymbol{memory}>$$<\boldsymbol{/memory}>$ or the agent calling result, you can answer the user’s question:
    \begin{itemize}
      \item $<\boldsymbol{think}> $
      The reason why you can answer the user's problem based on the current known information.$<\boldsymbol{/think}> $
      \item The answer to the current user's question.
    \end{itemize}

  \end{itemize}
\end{itemize}

\end{tcolorbox}

\begin{tcolorbox}[colframe=black, colback=gray!10, title=Template \#2: System Prompt for the E-commerce Function-call Agent, breakable]

  ~

  \textbf{Role:}

  ~

  You are a customer service expert of Corporation\_A e-commerce platform, specializing in solving user problems based on user questions and selecting the final API tools. Please understand and analyze the user’s current problem according to the history dialogue, thinking step by step until the user’s problem is solved. There are some tools available for selection at each step. You can think as follows:

  \begin{itemize}
    \item If the user’s intention can completely match other known specific APIs except for tool\_intention and tool\_retrievals, directly call that API and identify the required parameters.
    \item If the user’s intention is clear but cannot match other known specific APIs except for tool\_intention and tool\_retrievals, call tool\_retrievals to recall some related APIs to choose again.
    \item If an API has been called and returned results, or content has been recalled between $<\boldsymbol{memory}>$$<\boldsymbol{/memory}>$, you need to judge whether the current API result or the similar answers recorded in the memory meet the user’s intention. If so, directly output the answer based on it; otherwise, re-arrange.
  \end{itemize}

  ~

  \textbf{Specialized Retrieval Tools / Memory:}

  ~

  $<\boldsymbol{tools}>$

  \begin{itemize}
    \item \{"name": "tool\_retrievals", "description": "Knowledge retrieval tool. Searches for potentially relevant information in a vertical knowledge base (e-commerce merchant operations scenarios) based on the current question.", "arguments": \{\}\}
    \item ...
  \end{itemize}

  $<\boldsymbol{/tools}> $

  $<\boldsymbol{memory}> $

  {memory\_append}(Optional)

  $<\boldsymbol{/memory}> $

  ~

  \textbf{Problem Resolution Framework:}

  ~

  Specific requirements are as follows:

  \begin{itemize}
    \item When selecting tools, please refer to the tool’s function description. Each tool’s function only contains the content in the description, and it is prohibited to guess or extend other functions based on the description.
    \item When selecting APIs, you can only choose from the candidate API set recalled through tool\_retrievals. Each API’s function only contains the content in the description.
    \item Please be faithful to the semantics of the current problem and historical dialogue, and do not output content that does not exist in the historical dialogue and current problem.
    \item Do not call the same tool more than once, and try to call different APIs.
    \item Just output the tool response directly if there is no error.
    \item Output strictly according to the following format:

    \begin{itemize}
      \item The APIs in $<\boldsymbol{tools}>$$<\boldsymbol{/tools}>$ and all the currently recalled APIs cannot meet the user’s intention:
      $<\boldsymbol{think}>$
      The analysis and thinking process of the user’s problem, and the reason for calling the API retrieval tool.
      $<\boldsymbol{/think}>$
      $<\boldsymbol{tool\_call}>$
      \{"name": "tool\_retrievals", "arguments": \{"intention": user\_intention\}\}
      $<\boldsymbol{/tool\_call}>$

      \item The APIs in $<\boldsymbol{tools}>$$<\boldsymbol{/tools}>$ or all the currently recalled APIs can meet the user’s intention:
      $<\boldsymbol{think}>$
      The analysis and thinking process of the user’s problem, and the reason for choosing which api to solve the user's problem.
      $<\boldsymbol{/think}>$
      $<\boldsymbol{tool\_call}>$
      \{"name": chosen api name, "arguments": parameters passed to api\}
      $<\boldsymbol{/tool\_call}>$

      \item According to the recalled information between  $<\boldsymbol{memory}>$$<\boldsymbol{/memory}>$ or the API calling result, you can answer the user’s question:

      \begin{itemize}
        \item $<\boldsymbol{think}> $
        The reason why you can answer the user's problem based on the current known information.$<\boldsymbol{/think}> $
        \item The answer to the current user's question.
      \end{itemize}

    \end{itemize}

  \end{itemize}

  \end{tcolorbox}

\begin{tcolorbox}[colframe=black, colback=gray!10, title=Template \#3: System Prompt for the General Function-call Agent, breakable]

  ~

  \textbf{Role:}

  ~

  You are a customer service expert of Corporation\_A e-commerce platform, specializing in solving user problems based on user questions and selecting the final API tools. Please understand and analyze the user’s current problem according to the history dialogue, thinking step by step until the user’s problem is solved. There are some tools available for selection at each step. You can think as follows:

  \begin{itemize}
    \item If the user’s intention can completely match other known specific APIs except for tool\_intention and tool\_retrievals, directly call that API and identify the required parameters.
    \item If the user's intention is clear but cannot match other known specific APIs except for tool\_intention and tool\_retrievals, call tool\_retrievals to recall some related APIs to choose again.
    \item If an API has been called and returned results, or content has been recalled between $<\boldsymbol{memory}>$$<\boldsymbol{/memory}>$, you need to judge whether the current API result or the similar answers recorded in the memory meet the user’s intention. If so, directly output the answer based on it; otherwise, re-arrange.
  \end{itemize}

  ~

  \textbf{Specialized Retrieval Tools / Memory:}

  ~

  $<\boldsymbol{tools}>$

  \begin{itemize}
    \item \{"name": "tool\_retrievals", "description": "The API retrieval tool. Retrieve related APIs from the API knowledge base based on intention.", "arguments": \{"intention": "user's current intention"\}\}
    \item ...
  \end{itemize}

  $<\boldsymbol{/tools}> $

  $<\boldsymbol{memory}> $

  {memory\_append}(Optional)

  $<\boldsymbol{/memory}> $

  ~

  \textbf{Problem Resolution Framework:}

  ~

  Specific requirements are as follows:

  \begin{itemize}
    \item When selecting tools, please refer to the tool’s function description. Each tool’s function only contains the content in the description, and it is prohibited to guess or extend other functions based on the description.
    \item When selecting APIs, you can only choose from the candidate API set recalled through tool\_retrievals. Each API's function only contains the content in the description.
    \item Please be faithful to the semantics of the current problem and historical dialogue, and do not output content that does not exist in the historical dialogue and current problem.
    \item Do not call the same api more than once, and try to call different APIs.
    \item Just output the tool response directly if there is no error, or there is no other appropriate api to call.
    \item Output strictly according to the following format:

    \begin{itemize}
      \item The APIs in $<\boldsymbol{tools}>$$<\boldsymbol{/tools}>$ and all the currently recalled APIs cannot meet the user’s intention:
      $<\boldsymbol{think}>$
      The analysis and thinking process of the user’s problem, and the reason for calling the API retrieval tool.
      $<\boldsymbol{/think}>$
      $<\boldsymbol{tool\_call}>$
      \{"name": "tool\_retrievals", "arguments": \{"intention": user\_intention\}\}
      $<\boldsymbol{/tool\_call}>$

      \item The APIs in $<\boldsymbol{tools}>$$<\boldsymbol{/tools}>$ or all the currently recalled APIs can meet the user’s intention:
      $<\boldsymbol{think}>$
      The analysis and thinking process of the user’s problem, and the reason for choosing which api to solve the user's problem.
      $<\boldsymbol{/think}>$
      $<\boldsymbol{tool\_call}>$
      \{"name": chosen api name, "arguments": parameters passed to api\}
      $<\boldsymbol{/tool\_call}>$

      \item According to the recalled information between  $<\boldsymbol{memory}>$$<\boldsymbol{/memory}>$ or the API calling result, you can answer the user’s question:

      \begin{itemize}
        \item $<\boldsymbol{think}> $
        The reason why you can answer the user's problem based on the current known information.$<\boldsymbol{/think}> $
        \item The answer to the current user's question.
      \end{itemize}

    \end{itemize}

  \end{itemize}

  \end{tcolorbox}

  \begin{tcolorbox}[colframe=black, colback=gray!10, title=Template \#4: System Prompt for the QA Agent, breakable]

    ~

    \textbf{Role:}

    ~

    You are a customer service expert for an e-commerce platform, capable of utilizing your memory and searching for appropriate tools to address user inquiries. Based on the current question and past tool selections and their responses, proceed step-by-step to determine which tool to use or what content to output next.

    ~

    \textbf{Specialized Retrieval Tools / Memory:}

    ~

    1. Tools at your disposal: Results from the same tool \& arguments are unique.

    $<\boldsymbol{tools}> $

    \begin{itemize}
      \item \{"name": "tool\_retrievals", "description": "Knowledge retrieval tool. Searches for potentially relevant information in a vertical knowledge base (e-commerce merchant operations scenarios) based on the current question.", "arguments": \{\}\}
      \item \{"name": "tool\_summary", "description": "Intelligent Q\&A tool. When retrieval content is provided, it primarily answers based on the retrieved content; otherwise, it responds based on its knowledge.", "arguments": \{\}\}
    \end{itemize}

    $<\boldsymbol{/tools}> $

    2. Your memory content: Memory varies for different questions. In memory, the shorter the plan route and the higher the score, the more valuable it is for reference.

    $<\boldsymbol{memory}> $

    {memory\_append}(Optional)

    $<\boldsymbol{/memory}> $

    ~

    \textbf{Problem Resolution Framework:}

    ~

    \begin{itemize}
      \item Requirements for Tool Selection:

      \begin{itemize}
        \item tool\_retrievals: Call this tool when retrieving domain-specific knowledge related to merchants or e-commerce scenarios is required.
        \item tool\_summary: Call this tool when generating a response to the user's question is needed.

        \begin{itemize}
          \item tool\_summary will refer to the results from tool\_retrievals to generate a response only when the previous call was to tool\_retrievals; otherwise, it will respond directly.
        \end{itemize}

        \item Outputting Answers: (Must Pay Attention!) You cannot answer questions directly. The following scenarios apply when outputting answers:
        \begin{itemize}
          \item If a similar question exists in the memory, you can directly output the answer provided in the memory without invoking any tools.
          \item If no usable answer is found in the memory, you must first call tool\_summary and return its output result as is (without modifying the output of tool\_summary).
        \end{itemize}

        \item Tool Efficiency: Tool invocations incur costs. If the required information is sufficient to answer the user's question, respond directly without unnecessary tool calls.
      \end{itemize}

      \item Special Handling: When the question contains content such as “Pass calling ... Results are as follows ...”, this part represents the results of historical API calls. You do not need to answer this part in your response. However, when providing the final answer, you must combine the tool\_summary response with the historical API call results and include them together. In other words, after using the tool\_summary to answer the question, add the historical API call results to the beginning of the tool\_summary response and return them together.

      \item Applying the Above Tool Selection Requirements:
      When selecting tools for the current round, consider the following:

      \begin{itemize}
        \item Analyze Previously Called Tools:
        \begin{itemize}
          \item Important: Avoid calling the same tool that has already been called in history.
        \end{itemize}
        \item (If memory is not empty) Analyze memory content:
        \begin{itemize}
          \item If the user's question is essentially identical to one in memory, output the answer from memory without calling other tools.
          \item If the user's question is similar in content or type to one in memory, refer to the plan in memory for guidance. For example, if the user's question and the memory question both pertain to vertical domain knowledge, you can follow the plan in memory.
          \item If the user's question bears no similarity to the memory content, ignore the memory.
        \end{itemize}
        \item (After confirming memory does not provide a direct answer) Determine whether the user's question requires e-commerce or Corporation\_A-specific knowledge to answer:
        \begin{itemize}
          \item If no vertical domain knowledge is needed, call tool\_summary directly. Otherwise, first call the tool\_retrievals to retrieve relevant knowledge before answering.
        \end{itemize}
        \item If the Previous Call Was to tool\_summary:
        \begin{itemize}
          \item Since tool\_summary cannot be called again, and the output must come from tool\_summary, directly output the answer.
        \end{itemize}
      \end{itemize}

      \item Output format requirements:
        \begin{itemize}
          \item Answer output format: $<\boldsymbol{think}>$ Thought process $<\boldsymbol{/think}>$ Output answer(from tool\_summary or memory; Important: Any API call content found in the question must be included verbatim in the final response.)
          \item Tool call format: $<\boldsymbol{think}>$ Thought process $<\boldsymbol{/think}>$  $<\boldsymbol{tool\_call}>$ \{"name": "tool\_name", "arguments": \{"param": "value"\}\}$<\boldsymbol{/tool\_call}>$
        \end{itemize}
    \end{itemize}

    \end{tcolorbox}

\begin{tcolorbox}[colframe=black, colback=gray!10, title=Template \#5: System Prompt for the Math Agent, breakable]

~

\textbf{Role:}

~

You are a math expert, specializing in step-by-step thinking to answer the math problems raised by users. Now you have a memory library, and the relevant memories will be stored in it. You can combine the content in the memory to answer questions. The specific thinking steps are as follows:

\begin{itemize}
  \item If there is an identical question in the memory, you can use the answer of that question to directly answer the current question.
  \item If all the questions in the memory are different from the current user’s question, you need to think and answer by yourself.
\end{itemize}

~

\textbf{Specialized Retrieval Tools / Memory:}

~

$<\boldsymbol{tools}>$

[{tool\_append}](Optional)

$<\boldsymbol{/tools}> $

$<\boldsymbol{memory}> $

{memory\_append}(Optional)

$<\boldsymbol{/memory}> $

~

\textbf{Problem Resolution Framework:}

~

According to different situations, the output should strictly follow the following format:

\begin{itemize}
  \item If there is an identical question in the memory:

  \begin{itemize}
    \item $<\boldsymbol{think}>$ The reason for choosing the answer to the identical question.$<\boldsymbol{/think}>$
    \item  The answer to the current question.
  \end{itemize}

  \item If there is no identical question in the memory:

  \begin{itemize}
    \item $<\boldsymbol{think}>$ The reason for not choosing a question from memory, and the steps of thinking about the current user’s question.$<\boldsymbol{/think}>$
    \item The answer to the current question.
  \end{itemize}

\end{itemize}

\end{tcolorbox}

\paragraph{System prompt for the single agent.}
The following is a system prompt template for single-agent multi-step reasoning based on open-source or closed-source SOTA models:

\begin{tcolorbox}[colframe=black, colback=gray!10, title=Template \#6: System Prompt for the Single Agent, breakable]

~

\textbf{Role:}

~

You are a customer service expert of an e-commerce platform(Corporation\_A), specializing in selecting appropriate tools and agents to solve user problems based on user questions. Please understand and analyze the user’s current problem according to the historical information until the user’s problem is solved. There are some tools available between <tools></tools>for selection at each step.

~

\textbf{Specialized Retrieval Tools:}

~

$<\boldsymbol{tools}> $

\begin{itemize}

\item \{"name": "tool\_retrievals\_knowledge", "description": "Vertical knowledge base search tool (e-commerce merchant operations context). Identifies relevant information based on user queries.", "arguments": \{"intention": "user's current intention or query"\}\}

\item \{"name": "tool\_retrievals\_API\_shop", "description": "E-commerce platform API lookup. Retrieves relevant APIs from the API knowledge base using intent analysis.", "arguments": \{"intention": "user's current intention or query"\}\}

\item \{"name": "tool\_retrievals\_API\_general", "description": "General API lookup. Retrieves relevant APIs from the API knowledge base using intent analysis.", "arguments": \{"intention": "user's current intention or query"\}\}

\end{itemize}

$<\boldsymbol{/tools}> $

~

\textbf{Problem Resolution Framework:}

~

1. Question Types \& Response Protocols:

You may encounter different types of questions. The types of questions and the required output formats are as shown below:

\begin{itemize}

  \item Math problems:
  \begin{itemize}
    \item Provide direct solutions to numerical queries.
    \item Output in the following format (Provide the numerical answer directly after $<\boldsymbol{/think}> $, without units or any irrelevant characters): $<\boldsymbol{think}> $...$<\boldsymbol{/think}> $Final numeric answer
  \end{itemize}

  \item API scheduling problems:
  \begin{itemize}
    \item The APIs are divided into e-commerce platform APIs and general APIs.
    \item When API tools are required: Use relevant tool\_retrievals to identify candidate APIs (original/paraphrased queries accepted).
    \item Output the API call results in the following format: $<\boldsymbol{tool\_call}> ${"name": "API\_name", "arguments": {"key1":["value11", "value12"], "key2":["value21", "value22"]...}}$<\boldsymbol{/tool\_call}> $
    \item Some solutions require sequential API calls, but you can just call only one API at each step. Use prior outputs as inputs for subsequent calls.
  \end{itemize}

  \item Q\&A problems:
  \begin{itemize}
    \item Engage directly in casual conversations (greetings/jokes/daily topics).
    \item For e-commerce policy queries: Invoke tool\_retrievals\_knowledge for domain knowledge. Respond based on retrieved content.
  \end{itemize}

\end{itemize}

2. Tool/API Selection Guidelines:

\begin{itemize}
  \item The results of the previous Tool/API call will be returned in the format $<\boldsymbol{tool\_response}> $...$<\boldsymbol{/tool\_response}> $.
  \item The response format for API dispatching results is: “Pass calling ... Results are as follows: ...”. This result should generally be output to the user as-is to indicate the content of the API call. Additionally, if multiple API calls are involved, all relevant API call results must be merged and presented together to the user.
  \item When you feel that the current information is insufficient to provide a final output, you can call different tool\_retrievals or APIs as additional input to arrive at the definitive answer.
  \item Efficiency is crucial - minimize Tool/API calls as much as possible while ensuring accuracy.
\end{itemize}

3. Output Format Requirements:

Note: You must adhere to the following output formats; otherwise, no results will be generated.

\begin{itemize}
  \item When you determine that additional Tool/API calls are needed (Tool call format: API/tool\_retrievals): $<\boldsymbol{think}> $Thought process$<\boldsymbol{/think}> $$<\boldsymbol{tool\_call}> $\{"name": "tool\_name", "arguments": \{"param": "value"/["value"]\}\}$<\boldsymbol{/tool\_call}> $
  \item When you believe the current conclusion is sufficient to return to the user: $<\boldsymbol{think}> $Thought process$<\boldsymbol{/think}> $Output answer(if math problems, output final numeric answer; If it is an API-related issue and does not involve multiple API calls, output the content from $<\boldsymbol{tool\_response}> $ exactly as it is.)
\end{itemize}

\end{tcolorbox}

\begin{figure}[h]
\centering
\includegraphics[width=1.0\textwidth]{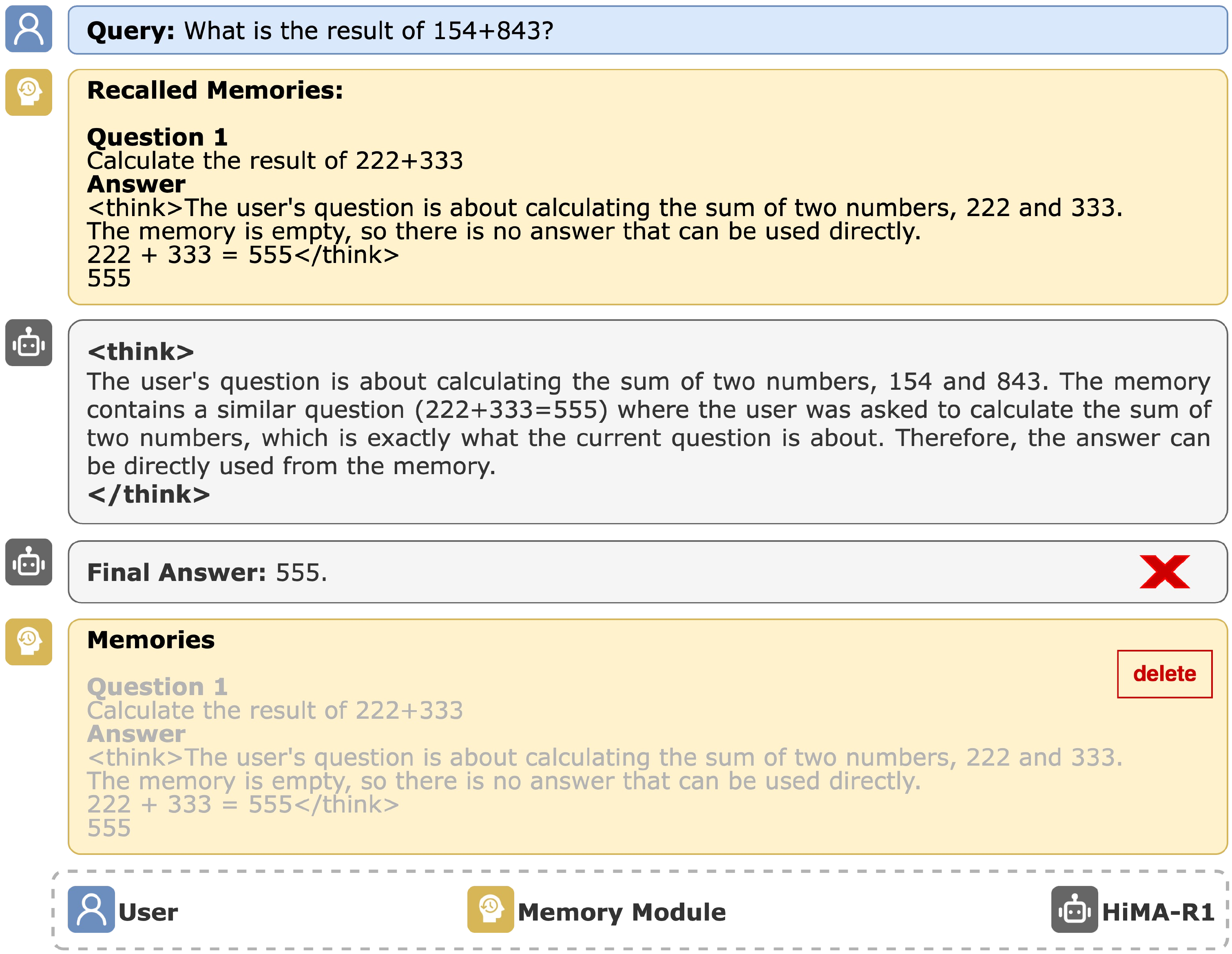}
\caption{An example of dynamic updating of the agent memory module.}
\vspace{-3mm}
\label{case_memory}
\end{figure}

\subsection{Case analyses}\label{Case}

\textbf{Dynamic update of the agent memory module.}
Figure~\ref{case_memory} illustrates the dynamic updating mechanism of the math-agent’s memory. When a user inputs an addition problem, the agent searches its memory repository and retrieves a similar problem whose numerical values differ but share the same structural pattern. However, due to interference from the recalled memory, the agent erroneously assumes the answer can be directly extracted from the memory without re-evaluating the new numerical values, resulting in an incorrect response. This erroneous memory entry is subsequently penalized through a scoring mechanism that reduces its retention probability in the memory buffer. Over successive iterations, such problematic memories are purged from the memory base, while memories that consistently contribute to accurate solutions are preserved. Eventually, the memory buffer converges to a stable state where it predominantly retains entries that enhance the agent’s problem-solving reliability.

\begin{figure}[!htbp]
\centering
\includegraphics[width=1.0\textwidth]{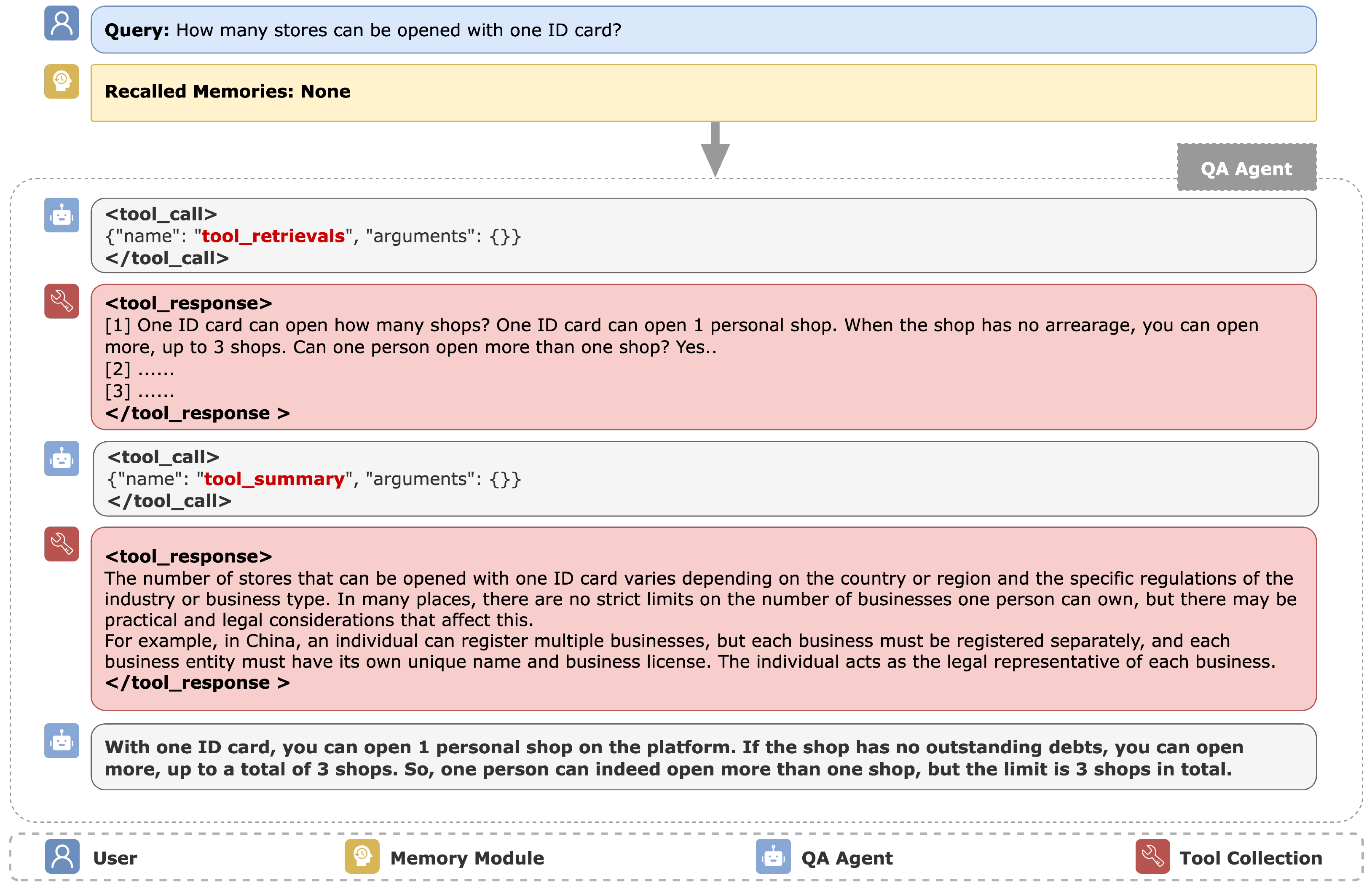}
\caption{The example of the QA agent in a question-answering task.}
\vspace{-3mm}
\label{case_qa}
\end{figure}

\begin{figure}[!htbp]
\centering
\includegraphics[width=1.0\textwidth]{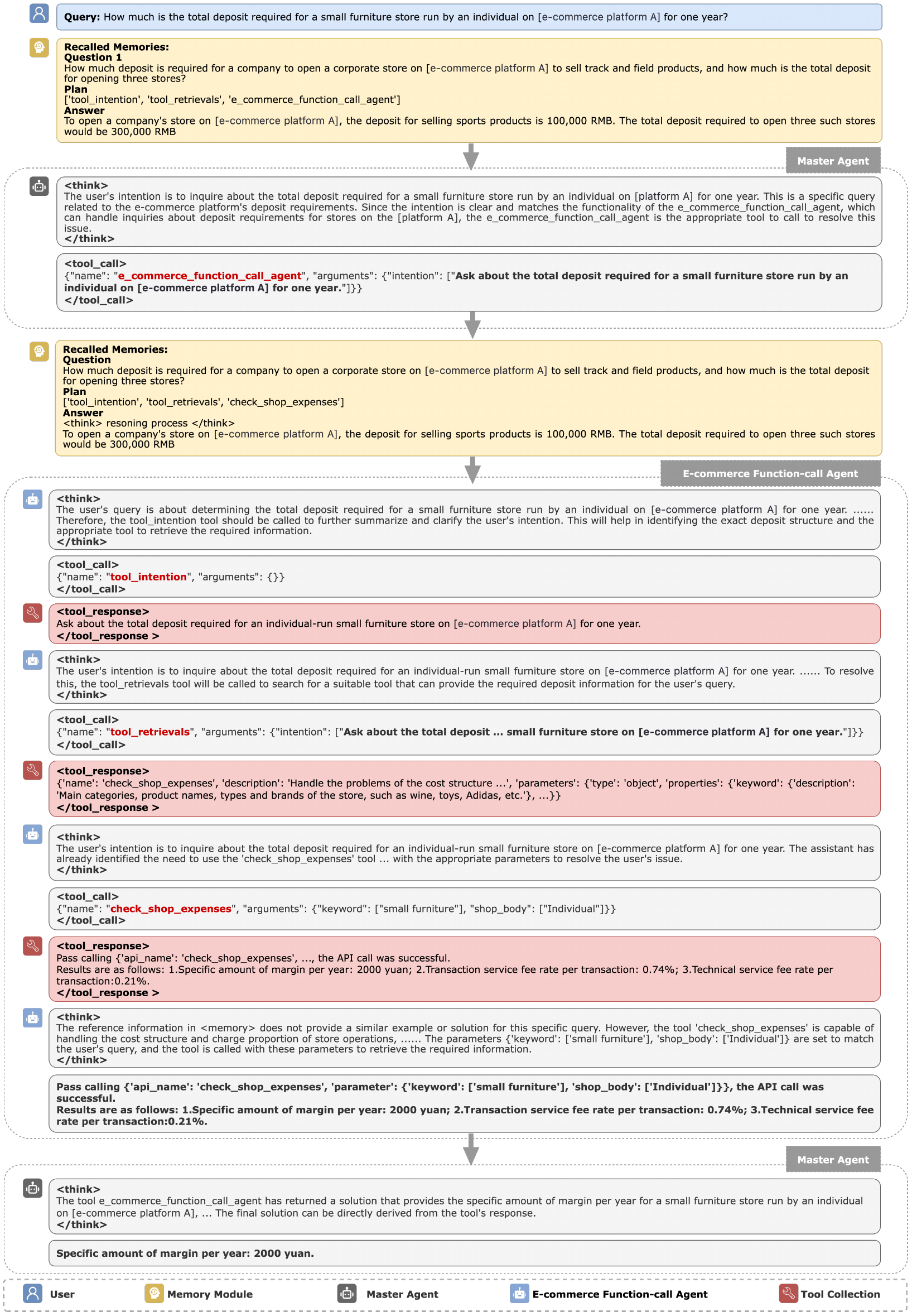}
\caption{The example of a complete reasoning trajectory in an e-commerce task.}
\vspace{-3mm}
\label{case_master}
\end{figure}

\textbf{More results for multiple tasks.}
Figure~\ref{case_qa} \textasciitilde Figure~\ref{case_master} depict the reasoning processes of agents across distinct tasks, illustrating the hierarchical architecture and modular interactions within the system. Figure~\ref{case_qa} outlines the QA Agent's reasoning process: upon receiving a user query, the master agent routes the task to the QA Agent, which first invokes the retrieval tool to fetch relevant knowledge entries from the external knowledge base and then calls the summary tool. In this case the summary drifts to a generic statement about business registration, and the agent answers from the retrieved platform rule instead, departing from Template \#4, which asks it to return the summary unchanged. The answer is then returned to the master agent for final delivery.
Figure~\ref{case_master} presents a complete reasoning process from query to response. When a user inquires about the annual guarantee deposit for opening a furniture store on an e-commerce platform, the master agent first queries its memory buffer and identifies a similar historical case. This triggers the invocation of the e-commerce function-call agent. Within this sub-agent, the memory buffer is again consulted and recalls a case about a corporate store, whose plan uses the intention tool, the retrieval tool, and the check\_shop\_expenses API. Since the recalled answer concerns a different store type, the agent does not reuse it and notes that memory holds no matching solution, yet its own calls follow the same tool sequence. It analyzes user intent, retrieves contextually relevant APIs, selects the check\_shop\_expenses API based on semantic matching, and executes it to obtain the deposit amount (\emph{e.g.}, "2000 yuan/year"). After validating the API response's relevance, the sub-agent returns the result to the master agent, which forwards it to the user. In summary, these cases underscore that task resolution necessitates coordination between the master agent and at least one specialized sub-agent, with memory buffers playing a pivotal role in both retrieving prior solutions and optimizing future responses through iterative learning.

\subsection{Usage of LLMs}\label{LLM}

Large Language Models (LLMs) are employed exclusively for grammar checking in this manuscript. They are not utilized for writing assistance, literature retrieval, or research ideation.

\clearpage

\end{document}